\documentclass[12pt]{article}

\usepackage{amsmath, amssymb, dsfont, mathrsfs, amsthm}
\usepackage{titling}
\usepackage{subfig}
\usepackage{enumitem}
\usepackage{algorithm}
\usepackage{algcompatible}
\usepackage{empheq} 
\usepackage{colonequals}
\usepackage{booktabs}
\usepackage{multirow} 
\usepackage{xcolor}
\usepackage[normalem]{ulem}
\usepackage[authoryear,round]{natbib}
\usepackage[
breaklinks = true,
colorlinks = true,
linkcolor = purple,
urlcolor  = black,
citecolor = purple,
anchorcolor = green,
hypertexnames = false
]{hyperref} 

\newtheorem{theorem}{Theorem}[section]
\newtheorem{proposition}[theorem]{Proposition}

\newtheorem{lemma}[theorem]{Lemma}

\newtheorem{definition}[theorem]{Definition}

\newcommand{\Bs}[1]{\boldsymbol{#1}}

\newcommand{\bx}{\mathbf{x}}

\newcommand{\bsX}{\boldsymbol{X}}

\newcommand{\bsalpha}{\boldsymbol{\alpha}}
\newcommand{\bsbeta}{\boldsymbol{\beta}}

\newcommand{\bstheta}{\boldsymbol{\theta}}

\newcommand{\bsPhi}{\boldsymbol{\Phi}}

\newcommand{\bsvarphi}{\boldsymbol{\varphi}}

\newcommand{\E}{\mathbb{E}}

\newcommand{\R}{\mathbb{R}}
	
\newcommand{\U}{\mathbb{U}}

\newcommand{\Tau}{\mathcal{T}}

\newcommand{\sD}{\mathscr{D}}

\newcommand{\hf}{\widehat{f}}

\DeclareMathSymbol{\sm}{\mathbin}{AMSa}{"39}
\DeclareMathOperator{\KL}{KL}

\DeclareMathOperator{\diag}{diag}
\DeclareMathOperator{\trace}{trace}

\newcommand{\Pxpipi}{\mathcal P_{\bx}(\Bs{\widehat\pi}, {\Bs{\pi}})}
\newcommand{\Pxpi}{\mathcal P_{\bx}(\Bs{\widehat\pi}, \cdot)}

\title{\bf Optimal Transport Aggregation for Distributed Mixture-of-Experts}
\author{Fa\"icel Chamroukhi$^{1}$,\, Nhat Thien Pham$^2$}
\date{
{\small $^1$IRT SystemX, 91120 Palaiseau, France\\
$^2$Normandie Univ, UNICAEN, CNRS, LMNO, 14000 Caen, France}\\
[0.5cm]
}
\usepackage{tikz}
\begin{document}

\maketitle

\renewcommand{\baselinestretch}{1.2}

\begin{abstract}
This paper considers the setting in which local Mixture-of-experts (MoE) models are fitted independently on distributed datasets and must be combined into a single global MoE model. Weighted averaging the local mixtures, whose experts are not aligned across machines, often yields a large mixture with many experts instead of $K$ and does not provide an estimator of the true model parameters. We propose to reduce this large mixture to a $K$-expert MoE by minimizing an expected transportation divergence, in which the transport plan depends on the covariates. A relaxation of this problem separates the experts from the gating network: the experts are computed by a majorization--minimization algorithm, which decreases the objective monotonically, and the gating network is then fitted to the gates implied by the optimal plan, with an explicit bound on the resulting loss. We show that the aggregated estimator is consistent whenever the local estimators are, without requiring the local models to have the same labeling across machines. The procedure only needs a single one-way communication round. Experiments on simulated and real data show accuracy close to centralized estimation at a fraction of its computation time.  Our source code is available at
\url{https://github.com/nhat-thien/Distributed-Mixture-Of-Experts}.
\end{abstract}

\noindent {\bf Keywords:} 
Mixture-of-Experts; 
Model Aggregation; 
Mixture Reduction; 
Optimal Transport; 
Majorization–Minimization;
One-Shot Federated Learning
{
\pagebreak
\renewcommand{\baselinestretch}{1.1}
\normalsize
\tableofcontents
\pagebreak
}

\renewcommand{\baselinestretch}{1.2}
\normalsize

\section{Introduction}

Modern machine learning applications increasingly involve datasets that are distributed across multiple machines.
A large body of work in distributed learning focuses on collaboratively optimizing model parameters across machines through iterative communication between nodes.
Typical approaches include distributed stochastic gradient descent and related optimization algorithms \citep{Zinkevich2010}.
While effective, these strategies often require frequent communication between machines, which can become a significant bottleneck in large-scale systems.

An alternative paradigm consists in performing standard inference locally and subsequently aggregating the resulting local estimators into a global model. This aggregation strategy allows local computations to be performed independently and reduces communications.
Several approaches following this divide-and-conquer principle have been proposed for classical models, including regression \citep{Mingxian,Chen2014ASA}, logistic regression \citep{Shofiyah2018}, clustering \citep{Zhao2009,NEURIPS2018_2fe5a27c}, and mixture models \citep{JMLR-distGMM}.

This paper focuses on parametric mixture-of-experts (MoE) \citep{jacobsME,jordan_and_xu_1995} with interpretable expert components.
They represent the conditional distribution of a response variable as a mixture of expert predictors, whose contributions depend on the input through a gating network.
This structure allows the model to capture nonlinear relationships and heterogeneous data while maintaining a probabilistic interpretation.
For an overview of theoretical and practical aspects of MoE modeling, we refer the reader to \citep{YukselWG12,NguyenChamroukhi-MoE}.

However, training MoE models on large-scale or decentralized datasets remains challenging. Distributed MoE systems have been studied in prior work \cite{YuLiang2024,song2025}, but these studies do not address global statistical estimation.
Standard training procedures rely on maximum likelihood estimation using the expectation-maximization (EM) algorithm \citep{jacobsME,jordan_and_xu_1995,dlr}.
These procedures become expensive when $N$ is large or when the data are distributed, so a natural strategy is to fit local MoE models and then combine them.  Aggregating MoE models, however, is not straightforward.
Simple averaging of local models may produce a model with an incorrect number of experts and does not preserve the structure of the MoE model.  Consequently, such standard aggregation strategies do not yield meaningful estimates of the parameters of the global model and may lead to models that are difficult to interpret.

To address this challenge, we propose an aggregation framework for MoE models based on optimal transport (OT).
It constructs a reduced global MoE estimator, with the correct number of experts, by minimizing an expected transportation divergence between the collection of local estimators and the aggregated model; the resulting problem is solved by a majorization--minimization (MM) algorithm.
The approach requires only a single round of unidirectional communication of model parameters 
and supporting covariates.
We show that the aggregated estimator is well-posed and consistent whenever the local estimators are consistent.

\paragraph*{Related work.}
Federated learning \citep{FLYang2019,federated-learning-monograph-2021} usually trains a shared model through many communication rounds, as in FedAvg \citep{mcmahan2017fedavg} or FedProx \citep{li2020fedprox} for heterogeneous data. Our setting is cross-silo with a single round, and the target is a parameter estimator rather than a predictor, so these methods are not directly comparable in terms of communication; they can however be applied to MoE models with gradient steps on the local log-likelihoods, and we use FedAvg with several rounds as a reference in Section~\ref{section: experimental study}. One-shot methods combine models trained separately \citep{guha2019oneshot}. For neural networks, hidden units are matched before averaging, with a Bayesian nonparametric model \citep{yurochkin2019pfnm,wang2020fedma} or with OT \citep{singh2020otfusion}. Local MoE models have the same label ambiguity for their experts, but here the transport is between the components of conditional mixtures and not between parameter vectors, and the target is an estimator of the model parameters. In this respect our setting is closer to divide-and-conquer estimation \citep{zhang2013commeff,ZhangDivideKernel2015,jordan2019commeff}. Many reduction algorithms exist for unconditional Gaussian mixtures \citep{crouse2011look}, for example greedy merging of pairs of components based on the KL divergence \citep{runnalls2007kullback}, or the minimization of a transportation divergence \citep{JMLR-distGMM}. Parameter estimation in MoE models is itself delicate: the convergence rates of the MLE depend on the form of the gating and expert functions and on their interaction \citep{ho2022convergence,nguyen2023demystifying}. Our result only uses the consistency of the local estimators and gives no rate; combining it with these results is left for future work.

Unlike the finite Gaussian mixtures of \citep{JMLR-distGMM}, MoE models have covariate-dependent gating functions: both the mixing proportions and the expert distributions depend on the inputs, which prevents the direct use of existing mixture reduction techniques.
 More precisely, the differences with \citet{JMLR-distGMM} are the following. The transport plan is a function of $\bx$ and the divergence is an expectation over $\bx$. The relaxation of Proposition~\ref{prop: infT infR} removes the gating network from the MM iterations; the gates are recovered from the optimal plan and then fitted by a softmax network, and Proposition~\ref{prop: gating projection} bounds the effect of this last step. Finally, the consistency result covers both the expert and the gating parameters.
\subsection{Mixture-of-Experts (MoE) context}

%
 A MoE model defines the conditional distribution of the response $y$ given the
  augmented  covariates vector $\bx\leftarrow(1,\bx^\top)^\top$
 as a mixture distribution with covariate-dependent mixing proportions and expert components:
\begin{equation}\label{eq: ME definition}
    f(y|\bx;\bstheta) \colonequals \sum_{k=1}^K \pi_k(\bx;\bsalpha)\,\varphi(y|\bx;\bsbeta_k),
\end{equation}
where $\bstheta=(\bsalpha^\top,\bsbeta_1^\top,\ldots,\bsbeta_K^\top)^\top$ denotes the full parameter vector, $K$ is the number of experts, $\bsbeta_k$ is the parameter of expert $k$.
The gating function $\pi_k(\bx;\bsalpha)$ is typically a softmax
where $\bsalpha=(\bsalpha_1^\top,\ldots,\bsalpha_{K-1}^\top)^\top$ and $\bsalpha_k\in\mathbb{R}^{d+1}$.
The expert components $\varphi(y|\bx;\bsbeta_k)$ may take different forms depending on the considered problem: Gaussian experts are common in regression \citep{jacobsME,nguyen2023demystifying} and multinomial logistic experts in classification \citep{nguyen2024multinomial}.
The class of MoE models is known to be dense and enjoys strong approximation capabilities \citep{nguyen2020approximationMoE}.

\subsection{Main contributions}

The main contributions of this work can be summarized as follows:

\begin{enumerate}[label=\roman*)]
\item
We introduce a distributed learning framework for mixture-of-experts models in which MoE estimators trained on decentralized datasets are aggregated into a single MoE estimator.

\item
We propose an aggregation strategy based on OT that constructs a reduction estimator for the global MoE model by minimizing a transportation divergence between the collection of local estimators and the aggregated model, 
preserving the structure of the original MoE model.

\item We derive a majorization--minimization (MM) algorithm, with a monotone descent property, for computing the aggregated estimator.

\item We establish theoretical guarantees for the proposed estimator: well-posedness and consistency of the empirical reduction estimator (Definition~\ref{def: reduction estimator}) under standard assumptions on the local estimators. The MM algorithm used to compute it decreases the objective monotonically and, under exact minimization of the empirical surrogate, its assignments stabilize after finitely many iterations.


\end{enumerate}

\section{Aggregating distributed mixture-of-experts}
\label{section: Aggregating ME models}

\subsection{Problem setting}
We consider the problem of one-shot aggregation of MoE models trained locally on decentralized datasets.
The goal is to combine the resulting local models at a central server into a reduced global MoE model.

Suppose that $M$ local machines own $M$ local datasets $\mathscr{D}_1,\ldots,\mathscr{D}_M$, respectively. Denote $N_m:=\vert\sD_m\vert$, and $N:=\sum_{m=1}^M N_m$, and $\mathscr{D} := \bigcup_{m=1}^M \sD_m$.
%
Note that when data are distributed for computational reasons,  many strategies can be adopted, including by randomly dividing them into enough small disjoint subsets or by constructing small bootstrap sub-samples like in \citep{BLB-JRSS}.

Let $f(\cdot|\bx,\bstheta^*)$ be the true unknown global MoE model of the data that taking the form in \eqref{eq: ME definition}, and $\bstheta^*$ be the true unknown parameter vector. For convenience, let us denote $f^* : =f(\cdot|\bx,\bstheta^*)$.

  Let $\hf_m$ be the local MoE fitted on $\sD_m$, e.g., via EM algorithm, with $K_m$ experts:
\begin{equation}\label{eq: local density}
    \hf_m \colonequals f(\cdot|\bx,\widehat\bstheta_m) = \sum_{k=1}^{K_m} \pi_k(\bx; \widehat\bsalpha^{(m)})
    \varphi(\cdot| \bx;\widehat\bsbeta_k^{(m)}),
\end{equation} 
where $\widehat\bstheta_m=(\widehat\bsalpha^{(m)},\widehat\bsbeta_1^{(m)},\ldots,\widehat\bsbeta_{K_m}^{(m)})$
is the locally estimated parameter vector. Let $L := \sum_{m=1}^M K_m$.
 %
The question is how to approximate the true global density $f^*$ from the local densities $\hf_m$, and more importantly, how to aggregate the local estimators $\widehat\bstheta_m$, with possibly different dimensions, to produce a single aggregated estimator for the true parameter vector $\bstheta^*$.
Hereafter, we use ``local density'' and ``local estimator'' to respectively refer to
$\hf_m$ and
$\widehat\bstheta_m$.
%
\subsection{Key challenges in aggregating distributed MoE models}
\label{sec: core technical}
A natural and intuitive estimator of $f^*$ is the weighted average
\begin{equation}\label{eq: weighted average f}
\bar f^W = \sum_{m=1}^M \lambda_m \hf_m,
\end{equation}
where $\lambda_m:=\frac{N_m}{N}$. In fact, it is a good approximation of $f^*$ since it is consistent when the local estimators are (see Step~1 of the proof of Theorem~\ref{thm: consistency}). However, $\bar f^W$ has two issues.
Firstly, it only provides an approximation to the conditional density value, i.e., $\bar f^W(y|\bx) \approx f^*(y|\bx)$ for each $\bx\in\mathcal X$,
but does not directly approximate the true parameter $\bstheta^*$, which is often the quantity of interest as it gives us a compact model with $K$ experts for prediction and clustering, and further analyses, for instance to compare experts across sites.
Secondly, if we substitute \eqref{eq: local density} into \eqref{eq: weighted average f}:
%
\begin{equation}\label{f^W full}
\bar f^W  = \sum_{m=1}^M \sum_{k=1}^{K_m} \lambda_m \pi_k(\bx; \widehat\bsalpha^{(m)}) \,\varphi(\cdot| \bx;\widehat\bsbeta_k^{(m)}),
\end{equation}
then we can see that $\bar f^W$ can be viewed as a MoE model with $L$ components.
Although this $L$-component MoE approximates well $f^*$, it has a too many components, i.e., $L$ instead of $K$, which makes the approximation results 
difficult to interpret.
Finally, even when all the local MoEs were fitted with $K_m=K$ for all machines, $\sum_{m=1}^{M} \lambda_m \pi_k(\bx; \widehat\bsalpha^{(m)})$ often fails
to approximate the true proportion $\pi^*_k(\bx; \bsalpha^*)$ due to label switching (Figure~\ref{Figure: toy illustration} in Appendix \ref{Example} illustrates this on a small example). Hence, this aggregating strategy is less desirable.

\subsection{The proposed aggregation strategy}\label{proposed aggregation strategy}

Let $\mathcal M_K$ denote the space of all $K$-component MoE models {as in \eqref{eq: ME definition}}.
When the gates are not required to be softmax functions we write $\widetilde{\mathcal M}_K$ (Proposition~\ref{prop: infT infR}).
Two other common approaches for aggregating the local densities $\hf_1, \ldots, \hf_M$ are via
\begin{align}
\bar f^B &\colonequals f(y|\bx;\bar\bstheta^B)= \underset{g\in\mathcal M_K}{\arg\inf}\sum_{m=1}^M \lambda_m  \rho(\hf_m,g),\label{eq: barycenter approx}\\
\bar f^R &\colonequals f(y|\bx;\bar\bstheta^R) = \underset{g\in\mathcal M_K}{\arg\inf}\ \rho(\bar f^W,g),\label{eq: reduction approx}
\end{align}where  $\rho$ is some divergence defined on the space of finite MoE distributions.
The solutions $\bar f^B$ and $\bar f^R$ are often known as barycenter and reduction solutions, respectively.
When $\rho$ is a Wasserstein distance, $\bar f^B$ is a Wasserstein barycenter \citep{agueh2011barycenters} restricted to $\mathcal M_K$, whereas $\bar f^R$ is a projection of the average $\bar f^W$ onto $\mathcal M_K$.

Both solve the wrong number of components in $\bar f^W$, and coincide for some choices of $\rho$. We use the reduction solution $\bar f^R$: it looks for the $K$-component model closest to $\bar f^W$, which already approximates $f^*$ well; it is cheaper to compute than \eqref{eq: barycenter approx}; and the barycenter can be counter-intuitive for some $\rho$, as shown in \citep{JMLR-distGMM} for univariate Gaussian mixtures under the $2$-Wasserstein divergence. Appendix~\ref{app: barycenter} gives the detailed comparison.

The proposed reduction strategy addresses these issues and provides a reduced MoE model $\bar f^R$ with the correct number of components, based on a well-posed and consistent reduction estimator $\bar{\boldsymbol{\theta}}^R$ of $\boldsymbol{\theta}^*$ whenever the local estimators are consistent.

\section{Estimator reduction via optimal transport}\label{section: algorithm}
We develop an algorithm to compute the reduction estimator $\bar\bstheta^R$.
We denote the components of $\bar\bstheta^R$ by
$\bar\bstheta^R = (\bar\bsalpha^R,\bar\bsbeta_1^R,\ldots,\bar\bsbeta_K^R)$.

By definition, the reduction estimator $\bar\bstheta^R$ in \eqref{eq: reduction approx} is the $K$-component MoE $g\in\mathcal M_K$ that is closest to the $L$-component MoE $\bar f^W$ w.r.t. the divergence $\rho$. In general, an analytic solution is difficult to obtain for such a reduction problem, especially in the context of MoE models where the conditional density is constructed upon several components.
Therefore, we have to find a numerical solution for $\bar f^R$ and $\bar\bstheta^R$.

%
Firstly, the solution depends strongly on the choice of the divergence $\rho$, which measures the goodness of a candidate $g\in\mathcal M_K$.
Best candidate minimizes its divergence from the large MoE model $\bar f^W$.
Secondly, computing the divergence between two MoE models is not explicit and is difficult. 
Borrowing the idea in \citep{JMLR-distGMM}, which finds a reduction estimator for finite Gaussian mixtures based on minimizing a transportation divergence between the large starting mixture and the desired mixture, we establish a framework for solving
\eqref{eq: reduction approx} in the subsequence.

\subsection{Expected transportation divergence for reduction estimator}
For notation convenience, we temporarily omit the parameters and rewrite $\bar f^W$ in \eqref{f^W full} simply as an $L$-component MoE as
\begin{align*}
\bar f^W = \sum_{\ell=1}^{L} \widehat\pi_\ell(\bx) \widehat\varphi_\ell(\cdot|\bx),
\end{align*}
where
$\left\{\widehat\pi_\ell(\bx)\right\}_{\ell\in[L]}$
stands for
$\big\{\lambda_m\pi_k(\bx;\widehat\bsalpha^{(m)})\big\}_{m\in[M]}^{k\in[K_m]}$
and
$\left\{\widehat\varphi_\ell(\cdot|\bx)\right\}_{\ell\in[L]}$
stands for
$\big\{\varphi(\cdot| \bx;\widehat\bsbeta_k^{(m)})\big\}_{m\in[M]}^{k\in[K_m]}$.
For $g\in\mathcal M_K$, we also write $g$ without parameters as
\begin{align}\label{eq: f and g}
    g = \sum_{k=1}^K \pi_k(\bx) \varphi_k(\cdot|\bx).
\end{align}

We need to define a divergence that measures the dissimilarity between the two MoE models $f^W$ and $g$.
Let $\Bs{\widehat\pi}$ and ${\Bs{\pi}}$ be, respectively, the column vectors of $\widehat\pi_\ell(\cdot)$'s and ${\pi}_k(\cdot)$'s.
For $\bx\in\mathcal X$, we denote
{\small\begin{align*}
    \Pxpipi
    &\colonequals \Bs\Pi\left(\Bs{\widehat\pi}(\bx), {\Bs{\pi}}(\bx)\right) \\
    &=
    \left\{
        \Bs{P} \in \mathbb U^{L\times K}\colon
        \Bs{P}\mathds{1}_K = \Bs{\widehat\pi}(\bx),
        \Bs{P}^\top\mathds{1}_L =  {\Bs{\pi}}(\bx)
    \right\},
 \end{align*}}where $\mathbb U:=[0,1]\subset \mathbb R$, $\mathds{1}_K$ and $\mathds{1}_L$ are vectors of all ones.
In other words, for $\bx\in\mathcal X$, $\Pxpipi$ denotes the set of all matrices $\Bs P$ of size $L\times K$, with entries $P_{\ell k}\in[0,1]$, satisfying the marginal constraints
{\small\begin{align}
    \sum_{k=1}^K P_{\ell k} = \widehat\pi_\ell(\bx),\ \forall\ell\in[L],
    \ \textrm{and}\
    \sum_{\ell=1}^L P_{\ell k} = \pi_k(\bx),\ \forall k\in[K].\label{eq: marginal constraints}
\end{align}}
Such a matrix $\Bs P \in \Pxpipi$ represents a transportation plan from the mixing proportion
$\Bs{\widehat\pi(\bx)}$ to the mixing proportion $\Bs\pi(\bx)$. The entry $P_{\ell k}$ indicates the amount of mass transported from
$\widehat\pi_{\ell}(\bx)$ to $\pi_k(\bx)$.

Let $\bsPhi=\{\varphi(\cdot|\bx;\bsbeta):\bsbeta\in\mathcal B\}$, the family of the component conditional densities $\varphi(\cdot|\bx;\bsbeta)$, e.g., the family of Gaussian regression experts, or logistic regression experts (for classification).

\begin{definition}[Expected transportation divergence]
The expected transportation divergence between two MoE models $h$ and $g$ is defined by
{\small\begin{align*}
    \Tau_{c}(h,g) &=
    \E\left[
    \underset{\Bs P\in\Pxpipi} {\inf}
    \sum_{\ell=1}^L \sum_{k=1}^K
        P_{\ell k}
        \ c\left(\widehat\varphi_\ell(\cdot|\bx),\varphi_k(\cdot|\bx)\right)
    \right] \nonumber \\
    &\equalscolon \E\Big[ \Tau_{c}(h,g,{\bx}) \Big],
\end{align*}}where $c(\cdot,\cdot)$ is a real-valued bivariate function defined on $\bsPhi\times\bsPhi$ that satisfies $c(\varphi_1,\varphi_2)\geqslant0$ for all $\varphi_1,\varphi_2\in\bsPhi$, and equality holds if and only if $\varphi_1=\varphi_2$.
\end{definition}
Here, the expectation is taken w.r.t. $\bx$, and $c$ is often referred to as cost function.
For a given,
$\Tau_{c}(h,g,{\bx})$ can be interpreted as the optimal total cost of transporting $\Bs{\widehat\pi(\bx)}$ to $\Bs\pi(\bx)$, where the unit transportation cost is proportional to the value $c\left(\widehat\varphi_\ell(\cdot|\bx),\varphi_k(\cdot|\bx)\right)$.
%
This interpretation is common in OT and is known as the Kantorovich formulation  \citep{Oberman2015AnEL}; see \citet{villani2009optimal,santambrogio2015optimal} for the theory and \citet{peyre2019computational} for computational aspects.
The value $\Tau_{c}(h,g)$ is therefore defined as the expectation of these optimal transportation costs.
%
%

By considering $\rho$ to be the expected transportation divergence $\Tau_{c}(\cdot,\cdot)$, problem \eqref{eq: reduction approx} becomes
{\small\begin{align}
\label{eq: minimize TD}
 \!\!\!\!   \bar f^R &=  \underset{g\in\mathcal M_K}{\arg\inf}\ \Tau_{c}(\bar f^W,g) \nonumber \\
    &= \underset{g\in\mathcal M_K}{\arg\inf}
        \ \E\left[
        \underset{\Bs P\in\Pxpipi} {\inf}
        \sum_{\ell=1}^{L} \sum_{k=1}^K
            P_{\ell k}
            \, c\left(\widehat\varphi_\ell(\cdot|\bx),{\varphi}_k(\cdot|\bx)\right)
        \right].
\end{align}}Problem \eqref{eq: minimize TD} involves two optimization problems: one over $\Pxpipi$ and one over $\mathcal M_K$.
%
%
%

We will show later that the constraint $\Bs P\in\Pxpipi$ can in fact be relaxed to $\Bs P \in \Pxpi$. This means that we only need $\Bs P$ to satisfy the constraints in the first part of \eqref{eq: marginal constraints}, those in the second part of \eqref{eq: marginal constraints} are redundant. In other words, instead of searching for a plan $\Bs P$ that satisfies $\Bs\pi(\bx)$, we can construct $\Bs\pi$ to match $\Bs P$, and that constructed $\Bs\pi$ forms that gating functions for $g$.

Let us define $\mathcal R_c(\bar f^W, g)$ as a function of $g$ as follows
{\small \begin{align}\label{eq: relaxed objective function}
    \mathcal R_c(\bar f^W, g)
    &= \E\left[
        \underset{\Bs P\in\Pxpi} {\inf}
        \sum_{\ell=1}^{L} \sum_{k=1}^K
            P_{\ell k}
            \ c\left(\widehat\varphi_\ell(\cdot|\bx),{\varphi}_k(\cdot|\bx)\right)
        \right] \nonumber \\
        &\equalscolon \E\Big[ \mathcal R_{c}(\bar f^W,g,{\bx}) \Big].
\end{align}}
Note that, different from $\Tau_{c}(\bar f^W,g)$, in $\mathcal R_c(\bar f^W, g)$, $\Bs P$ is free of the constraints involving the gating network $\Bs\pi$ of $g$.
The following proposition shows that minimizing $\Tau_{c}(\bar f^W,\cdot)$ over the enlarged class $\widetilde{\mathcal M}_K$ amounts to minimizing $\mathcal R_c(\bar f^W,\cdot)$, which is much easier and gives the same expert parameters; the induced gates are then projected onto the softmax family, with the resulting gap controlled by Proposition~\ref{prop: gating projection}. Before presenting the proposition, we define $\mathcal P(\bar f^W,g,\bx)$, a function of $g\in\mathcal M_K$ and $\bx\in\mathcal X$:
{\small\begin{equation}\label{eq: compute optimal plan}
    \mathcal P(\bar f^W,g,\bx) = \underset{\Bs P\in\Pxpi} {\arg\inf}
    \sum_{\ell=1}^{L} \sum_{k=1}^K
    P_{\ell k}
    \ c\left(\widehat\varphi_\ell(\cdot|\bx),{\varphi}_k(\cdot|\bx)\right).
\end{equation}}This function returns a minimizer of $\mathcal R_{c}(\bar f^W,g,{\bx})$.
For simplicity, let the dependence of the functions on $\bar f^W$ remains in the background, so that 
$\Tau_{c}(\bar f^W,g)$,
$\mathcal R_c(\bar f^W, g)$,
and $\mathcal P(\bar f^W,g,\bx)$
will be written as
$\Tau_{c}(g)$
$\mathcal R_c(g)$,
and $\mathcal P(g,\bx)$, respectively.

\begin{proposition}\label{prop: infT infR}
    Let $\widetilde{\mathcal M}_K$ be the class of $K$-component MoE models whose experts are in $\bsPhi$ and whose gating functions are arbitrary measurable maps from $\mathcal X$ to the probability simplex in $\R^{K}$. Then
          $\underset{g\in\widetilde{\mathcal M}_K}{\inf}\Tau_{c}(g)
         = \underset{g\in\widetilde{\mathcal M}_K}{\inf}\mathcal R_c(g).$
    The reduction solution in $\widetilde{\mathcal M}_K$ is hence given by $\widetilde f^R =  \underset{g\in\widetilde{\mathcal M}_K}{\arg\inf}\ \mathcal R_c(g)$
    and the gating functions of $\widetilde f^R$ can be calculated by
    \begin{equation}\label{eq: sum of gating network}
         {\pi}_k(\bx) = \sum_{\ell=1}^{L}\mathcal P_{\ell k}(\widetilde f^R, \bx),\quad\forall \bx\in\mathcal X,
    \end{equation}
    where $\mathcal P_{\ell k}(\widetilde f^R, \bx)$ denotes the entry $(\ell,k)$ of $\mathcal P(\widetilde f^R, \bx)$.
\end{proposition}
%
The proof of Proposition~\ref{prop: infT infR} is given in Sec.~\ref{appendix: proof infT infR}
of the supplementary material.
Since $\mathcal R_c$ does not depend on the gating functions of $g$ (they do not enter $\Pxpi$), minimizing it over $\widetilde{\mathcal M}_K$ or over $\mathcal M_K$ gives the same experts. The difference is in the gates: the functions in \eqref{eq: sum of gating network} are not softmax functions in general. We fit a softmax gating network to them (Section~\ref{subsec: MM algorithm}), and Proposition~\ref{prop: gating projection} bounds what is lost in $\Tau_c$. This leads to the following definition, which is the estimator computed by our algorithm and 
studied in Theorem~\ref{thm: consistency}.

\subsection{Construction of the reduction estimator}

As we can see, $\mathcal R_c(g)$ does not depend on the gates of $g$, so $\mathcal R_c(g)$ is actually  $\mathcal R_c(\Bs\varphi)$, where ${\Bs\varphi}=(\varphi_1,\ldots,\varphi_K)$ is the experts component of $g\in \widetilde{\mathcal M}_K$.
Thanks to Proposition~\ref{prop: infT infR}, we can construct the reduction estimator $\bar\bstheta^R=(\bar\bsalpha^R,\bar\bsbeta^R_1,\ldots,\bar\bsbeta^R_K)$ as follows.

\begin{definition}[Reduction estimator]\label{def: reduction estimator}
The targeted reduction estimator $\bar\bstheta^R=(\bar\bsalpha^R,\bar\bsbeta^R_1,\ldots,\bar\bsbeta^R_K)$ in \eqref{eq: reduction approx} is constructed in two stages:
(i) $\bar\bsbeta^R_k$ are the experts' parameters of $\bar g \in\widetilde{\mathcal M}_K$ obtained by $\bar{\bsvarphi}^R \in  \underset{\bsvarphi \in \bsPhi^K}{\arg\inf}\ \mathcal R_{c}(\bsvarphi)$:
\begin{align}\label{eq: problem minimize R}
    \bar{\bsvarphi}^R 
    \in \underset{\bsvarphi \in \bsPhi^K}{\arg\inf}
        \ \E\left[
        \underset{\Bs P\in\Pxpi} {\inf}
        \sum_{\ell=1}^{L} \sum_{k=1}^K
            P_{\ell k}
            \, c\left(\widehat\varphi_\ell(\cdot|\bx),{\varphi}_k(\cdot|\bx)\right)
        \right];
\end{align}(ii) With $a_k(\bx)=\sum_{\ell=1}^{L}\mathcal P_{\ell k}(\bar g,\bx)$,
\begin{align}\label{eq: find alpha}
\bar\bsalpha^R\in\arg\max_{\bsalpha}\ \E\Big[\sum_{k=1}^K a_k(\bx)\log\pi_k(\bx;\bsalpha)\Big].
\end{align}In practice, expectations are replaced by empirical means on the supporting sample $\bsX_S$. We write $\bar f^R=f(\cdot|\bx;\bar\bstheta^R)\in\mathcal M_K$ the reduction density, and refer to $\bar\bstheta^R$ as the OT-based reduction estimator (OT-RED).
\end{definition}
The optimization \eqref{eq: problem minimize R} can be done with the help of a Majorization-Minimization (MM) algorithm (e.g., \citep{lange2004mm}), while \eqref{eq: find alpha} can be solved efficiently using, e.g., a Newton--Raphson algorithm.
We defer the numerical approach for solving problem \eqref{eq: problem minimize R} to Section~\ref{subsec: MM algorithm}. We now show that the problem \eqref{eq: problem minimize R} is well-posed and state the conditions for the consistency of the reduction estimator.
In \eqref{eq: reduction approx}, $\bar f^R$ denoted a general solution; from now on it denotes the density of the reduction estimator in Definition~\ref{def: reduction estimator}.

\subsection{Well-posedness and consistency of the reduction estimator}

Let $\mathcal C(\Bs P, \bsvarphi)$ be a function of the transport plan w.r.t. the component densities of $g$, given by
 $$\mathcal C(\Bs P, \bsvarphi) = \sum_{\ell=1}^{L} \sum_{k=1}^K P_{\ell k} \ c\left(\widehat\varphi_\ell(\cdot|\bx),{\varphi_k}(\cdot|\bx)\right),$$ where $\Bs P\in\U^{L\times K}$, $\U=[0,1]\subset\R$, and $\bsvarphi = (\varphi_1(\cdot|\bx),\ldots,\varphi_K(\cdot|\bx)) \in\bsPhi^K$.
%
%
%
Let $\mathcal I(\bsvarphi,\bx)$ be a function of $\bsvarphi$ and $\bx$ defined by $\mathcal I(\bsvarphi,\bx)  = \inf_{\Bs P\in\Pxpi} \mathcal C(\Bs P, \bsvarphi),$  i.e., the optimal cost given $\bsvarphi$ and $\bx$ inside the expectation operator in \eqref{eq: problem minimize R}.
%
Then, the objective function of problem \eqref{eq: problem minimize R} can be written as $$\mathcal R_c(\bsvarphi) = \E\left[\mathcal I(\bsvarphi,\bx)\right].$$

Assumptions A1.--A4. require the true model to be identifiable with distinct experts, the cost to vanish only between equal experts, the covariate and parameter spaces to be compact with a continuous cost, and the covariates to admit a density; they are stated in Appendix~\ref{appendix: Assumptions}.

\begin{proposition}\label{prop: continuous and convex}
 Under Assumptions~\ref{assumption: continuous}--\ref{assumption: convex} (see Appendix \ref{appendix: Assumptions}), $\mathcal I(\bsvarphi,\bx)\leqslant\bar C$ for all $\bsvarphi\in\bsPhi^K$ and $\bx\in\mathcal X$, and $\mathcal R_c(\bsvarphi)$ is continuous on $\bsPhi^K$.
 %
 %
 It follows that the problem \eqref{eq: problem minimize R} has a global solution.
\end{proposition}
The proof of Proposition~\ref{prop: continuous and convex} is provided in Section~\ref{appendix: continuous and convex}
of the supplementary material.
The solution is not unique: permuting the experts does not change $\mathcal R_c$. Note also that $\mathcal R_c$ is not convex in general, since $\mathcal M_K$ is not a convex set and $\mathcal I(\bsvarphi,\bx)=\sum_{\ell}\widehat\pi_\ell(\bx)\min_{k} c(\widehat\varphi_\ell(\cdot|\bx),\varphi_k(\cdot|\bx))$ is a minimum of functions of $g$. Existence only uses continuity and compactness.

%
%

The following theorem establishes consistency of $\bar\bstheta^R$.

%

\begin{theorem}[Consistency of the reduction estimator]
\label{thm: consistency}
Let $\bar\bstheta^{R}$ be the reduction estimator of Definition~\ref{def: reduction estimator} computed on $\bsX_S$, where the experts minimize the empirical version of $\mathcal R_c$ up to $o_P(1)$.
Suppose that Assumptions~\ref{assumption: iid}--\ref{assumption: separation} hold, that $M$ is fixed with $N_m/N\to\lambda_m>0$,
and that each local estimator $\widehat\bstheta_m$ is a consistent estimator
of $\bstheta^*$ up to a relabeling of its components.
Then $\bar{\boldsymbol{\theta}}^{R}$ is a consistent estimator of $\boldsymbol{\theta}^*$ as $N,S\to\infty$.
\end{theorem}
The ordering and initialization assumption in Assumption~\ref{assumption: iid}
is standard in MoE models and ensures identifiability of the MoE model
\citep{JIANG19991253}. Assumption~\ref{assumption: continuous} is used to identify the true experts, and Assumptions~\ref{assumption: convex} and~\ref{assumption: separation} give the uniform continuity and the identifiability of the experts used in the proof. 
The proof of Theorem~\ref{thm: consistency} is provided in the supplementary
material (Section~\ref{appendix: consistency}).
The local models do not need to be aligned with each other: each of them may label its experts differently. This is where the weighted average $\bar\bstheta^W$ fails, and $\bar\bstheta^R$ is not affected by these labels.

\section{An MM algorithm for the reduction estimator}\label{subsec: MM algorithm}

%

We derive an MM algorithm \citep{lange2004mm}, which relies on the construction of a majorant function, to solve the problem \eqref{eq: problem minimize R}.
The MM algorithm in our setting proceeds as follows. Starting from an initial point $\bsvarphi^{(0)}\in\bsPhi^K$, at each iteration $t$ we construct a majorant function of $\mathcal R_c(\bsvarphi)$ at $\bsvarphi^{(t)}$, and minimize it to obtain the next iterate $\bsvarphi^{(t+1)}$. The generated sequence $(\bsvarphi^{(t)})_{t\geqslant1}$ satisfies the descent property
\(
\mathcal R_c(\bsvarphi^{(0)}) \geqslant \mathcal R_c(\bsvarphi^{(1)}) \geqslant \ldots
\)
The iterations are continued until convergence. On the supporting sample the algorithm stops after a finite number of iterations (Proposition~\ref{prop: MM termination}), and the experts of the last iterate give $\bar\bsbeta^R$ in Definition~\ref{def: reduction estimator}.

This descent property provides a stable optimization procedure, whereas monotonicity does not hold in general for gradient-based methods.

\subsection{The construction of majorant function}
The following provides a majorant for $\mathcal R_c(\bsvarphi)$.

\begin{proposition}\label{prop: majorant function}
 Let $\bsvarphi^{(t)}$ be the experts obtained at the $t$-th MM iteration and let
 {\small\begin{equation}\label{eq: update majorant function}
     \mathcal S_c(\bsvarphi,\bsvarphi^{(t)}) =
     \E \left[
     \sum_{\ell=1}^{L} \sum_{k=1}^K
     \mathcal P_{\ell k}(\bsvarphi^{(t)}, \bx)\ c(\widehat\varphi_\ell(\cdot|\bx), \varphi_{k}(\cdot|\bx))
     \right],
 \end{equation}}where $\mathcal P_{\ell k}(\bsvarphi^{(t)}, \bx)$ is given by
 {\small\begin{align}\label{eq: update transportation plan}
    \mathcal P_{\ell k}(\bsvarphi^{(t)}, \bx) =
    \left\{
        \begin{array}{ll}
            \widehat\pi_{\ell}(\bx) \!\! & \!\! \text{if } k = \underset{k^{\prime}\in[K]}{\arg\inf}\ c\big(\widehat\varphi_\ell(\cdot|\bx),\\
            \!\! & \!\! \hspace{2.2cm}\varphi_{k^\prime}^{(t)}(\cdot|\bx)\big),\\
            0          & \text{otherwise}.
        \end{array}
    \right.
\end{align}}Then $\mathcal S_c(\bsvarphi,\bsvarphi^{(t)})$ is a majorant of $\mathcal R_c(\bsvarphi)$ at $\bsvarphi^{(t)}$.
\end{proposition}
\footnote{In \eqref{eq: update transportation plan}, ties are broken by keeping the expert chosen at the previous iteration, and otherwise by taking the smallest index; the same rule defines $\mathcal P(\bar f^W,g,\bx)$ in \eqref{eq: compute optimal plan}, which makes $\bx\mapsto\mathcal P(\bar f^W,g,\bx)$ measurable. For Gaussian experts ties only occur on a $\mu$-null set.} The proof of Proposition~\ref{prop: majorant function} is provided in Section~\ref{appendix: majorant function}
of the supplementary material.
 Here we observe that the plan with entries $\mathcal P_{\ell k}(\bsvarphi^{(t)}, \bx)$ given in \eqref{eq: update transportation plan} is an optimal plan for the optimization problem \eqref{eq: compute optimal plan} when $\bsvarphi=\bsvarphi^{(t)}$ (this is clarified further in the proof of the proposition).

 The next step consists in minimizing $\mathcal S_c(\bsvarphi,\bsvarphi^{(t)})$ with respect to $\bsvarphi$ in order to obtain the next iterate $\bsvarphi^{(t+1)}$ of the MM algorithm. Observe that, the optimization of $\mathcal S_c(\bsvarphi,\bsvarphi^{(t)})$ can be carried out separately for each expert component. At iteration $t$, the parameter vector of expert $k$ can be updated by solving
 {\small
\begin{equation}\label{eq: update expert k}
    \varphi_k^{(t+1)}(\cdot|\bx)
    = \underset{\varphi\in\bsPhi}{\arg\inf}\,
    \E\left[
    \sum_{\ell=1}^{L}
    \mathcal P_{\ell k}(\bsvarphi^{(t)}, \bx)\, c(\widehat\varphi_\ell(\cdot|\bx), \varphi(\cdot|\bx))
    \right].
\end{equation}}According to the specification of the experts and the choice of the cost function $c$, problem \eqref{eq: update expert k} takes a particular form that can be solved efficiently, for example in the cases of Gaussian and logistic regression experts when $c$ is the KL-divergence: closed-form updates for Gaussian experts, and a convex weighted logistic regression, solved by Newton iterations, for logistic experts.
 Hence, the proposed algorithm alternates between the following two steps until convergence:
\begin{itemize}
    \item computing the majorant function $\mathcal S_c(\bsvarphi,\bsvarphi^{(t)})$ as in \eqref{eq: update majorant function}, which essentially amounts to updating $\mathcal P_{\ell k}(\bsvarphi^{(t)},\bx)$ according to \eqref{eq: update transportation plan};
    \item updating the expert parameters by solving \eqref{eq: update expert k}.
\end{itemize}The gating  parameters are updated using equation \eqref{eq: sum of gating network}, where the theoretical solution is replaced by the model $\bar \bsvarphi$ obtained by the MM algorithm (Definition~\ref{def: reduction estimator}). For this purpose, a supporting sample $\bsX_S$ available at the central server is required; details are provided in Section \ref{app: DETAILS OF UPDATING} of the supplementary material.
%
On the supporting sample, the assignment induced by \eqref{eq: update transportation plan} stabilises after finitely many iterations and the MM iterations reach a fixed point (Proposition~\ref{prop: MM termination}, proved in Section~\ref{appendix: MM termination} by the argument used for $k$-means). The fixed point need not be a global minimizer of $\mathcal R_c$.

The following result bounds the effect of replacing the gates \eqref{eq: sum of gating network} by softmax gates.
\begin{proposition}\label{prop: gating projection}
Let $\bar g$ be a MoE with experts $\bar\varphi_1,\ldots,\bar\varphi_K$, let $a_k(\bx)=\sum_{\ell=1}^{L}\mathcal P_{\ell k}(\bar g,\bx)$, and let $\bar f$ be the MoE with the same experts and gates $\pi_k(\bx;\bsalpha)$, for any $\bsalpha$. Then
\[
\Tau_c(\bar f)\leqslant\mathcal R_c(\bar g)+\tfrac12\,\E\big[\bar c(\bx)\,\|\Bs a(\bx)-\Bs\pi(\bx;\bsalpha)\|_1\big],
\]
where $\bar c(\bx)=\max_{\ell,k}c(\widehat\varphi_\ell(\cdot|\bx),\bar\varphi_k(\cdot|\bx))$.
If the experts $\bar\varphi_k$ minimize $\mathcal R_c$, then $\Tau_c(\bar f)-\inf_{g\in\widetilde{\mathcal M}_K}\Tau_c(g)$ is bounded by the same expectation. With the experts and gates of Definition~\ref{def: reduction estimator}, this applies to $\bar f=\bar f^R$.
\end{proposition}
The proof is in Section~\ref{appendix: gating projection}. The bound is zero when the gates \eqref{eq: sum of gating network} are exactly softmax functions, and it can be computed on $\bsX_S$ after the fit.

The gating family only enters this last step. Softmax gates can be replaced by other gating functions, such as Gaussians \citep{nguyen2024gaussiangated} or top-$K$ sparse softmax gates \citep{nguyen2024topk}, by changing the fit \eqref{eq: calculate gating network parameter}; Proposition~\ref{prop: gating projection} holds for any gating family.

\paragraph*{What is computed and what is proved.} Algorithm~\ref{algo: main algorithm} computes the estimator of Definition~\ref{def: reduction estimator}: experts from the MM algorithm, which decreases $\mathcal R_c$ and stops at a fixed point (Proposition~\ref{prop: MM termination}), then gates from the softmax fit. The gate fit does not change the experts, which remain a fixed point of the MM iterations. Theorem~\ref{thm: consistency} assumes that the experts are a global minimizer of the empirical objective; the MM algorithm does not guarantee it, which is why we use several initializations. The expectations are replaced by empirical means on $\bsX_S$, and the theorem is stated for $S\to\infty$. In this procedure the experts are computed without the gates, since $\mathcal R_c$ does not depend on them, and the interaction between gates and experts only enters through the fit of the gates. Minimizing $\Tau_c$ directly over $\mathcal M_K$, by alternating between the experts (with the plan constrained by the current gates) and the gating parameter, would take this interaction into account, we did not pursue it here.

\subsection{Communication and computation costs.}
Each machine sends $\widehat\bsalpha^{(m)}\in\R^{(K-1)(d+1)}$, the $K$ vectors $\widehat\bsbeta_k^{(m)}\in\R^{d+1}$, the $K$ variances and $N_m$, that is $(K-1)(d+1)+K(d+2)+1=\mathcal O(Kd)$ numbers. The communication cost for the parameters is therefore $\mathcal O(MKd)$, paid once. The aggregation also needs the covariates of the supporting sample (the responses are not used in the updates of Section~\ref{app: DETAILS OF UPDATING}), at a cost of $\mathcal O(Sd)$, with $S=N/M$ in our experiments. For comparison, $T$ rounds of FedAvg on the same model cost $\mathcal O(TMKd)$ in each direction.
The supporting sample is used only to approximate expectations in the aggregation step and does not affect the local training phase.
It can be obtained by randomly sampling a small subset of observations from the distributed datasets, or replaced by an auxiliary dataset already available at the server. Only covariates are needed. When sending covariates is not allowed, as in some federated settings, $\bsX_S$ can be replaced by public covariates from the same population, or by covariates simulated from a model of $\bx$ fitted on each machine; Table~\ref{tab:sensitivity_S} compares these choices.
Moreover, communication occurs only once
in contrast with many distributed optimization approaches where repeated communication rounds, often in both directions or between nodes, are required.
The proposed approach is therefore frugal in terms of communication, which is often a critical bottleneck in large-scale distributed learning.


\section{Explemental study}
\label{section: experimental study}

{
We evaluate the proposed OT-based reduction estimator $\bar\bstheta^R$ (OT-RED) against the centralized benchmark $\bstheta^G$ (GLB); the aggregation-based baselines: naive weighted average $\bar\bstheta^W$ (W-AVR), aligned average $\bar\bstheta^A$ (A-AVR), greedy merging $\bar\bstheta^{GM}$ (GM); the transportation medoid $\bar\bstheta^{MED}$ (MED); and the federated average $\bar\bstheta^{\mathrm{FED}}$ (FED) with number of rounds $T\in\{1, 5, 10, 20\}$.
The metrics include: $\mathcal T_c(\cdot, f^*)$, $\mathcal T_c(\cdot, \bar f^W)$, testing log-likelihood, $\text{MSE}(\cdot, \bstheta^*)$,  relative prediction error (RPE), adjusted Rand index (ARI), and learning time. Baselines and metrics are detailed in Appendix~\ref{Baselines and metrics}.

\subsection{Simulated data}
\begin{table*}[!ht]
\centering
\caption{Aggregation performance under uncoordinated local expert capacities ($M=16, K_m=3, L=48 \to K=5, d=20, N=100k$, mean $\pm$ std over 50 runs).}
\label{tab:official_heterogeneous}
\setlength{\tabcolsep}{5pt}
\resizebox{\textwidth}{!}{
\begin{tabular}{lccccccc}
\toprule
\textbf{Estimator} & $\mathcal{T}_c(\cdot,f^*)$ & $\mathcal{T}_c(\cdot,\bar f^W)$ & \textbf{Test Loglik} & $\textbf{MSE}(\cdot, \bstheta^*)$ & \textbf{RPE} & \textbf{ARI} & \textbf{Time (s)} \\
\midrule
GLB & $12.34 \pm 1.97$ & $105.31 \pm 4.88$ & $-52182.0 \pm 139.7$ & $0.0004 \pm 0.0001$ & $0.3917 \pm 0.0057$ & $0.9290 \pm 0.0024$ & $4255.698 \pm 283.415$ \\
\midrule
OT-RED + warm start (ours) & $13.47 \pm 2.21$ & $\mathbf{104.13 \pm 4.84}$ & $-55333.0 \pm 117.2$ & $\mathbf{0.0456 \pm 0.0008}$ & $\mathbf{0.4834 \pm 0.0049}$ & $0.9155 \pm 0.0031$ & $30.721 \pm 24.412$ \\
OT-RED + kmeans clustering (ours) & $13.47 \pm 2.21$ & $\mathbf{104.13 \pm 4.84}$ & $-55333.0 \pm 117.2$ & $\mathbf{0.0456 \pm 0.0008}$ & $\mathbf{0.4834 \pm 0.0049}$ & $0.9155 \pm 0.0031$ & $34.015 \pm 26.737$ \\
GM & $\mathbf{12.86 \pm 2.11}$ & $104.71 \pm 4.90$ & $\mathbf{-55332.5 \pm 117.8}$ & $\mathbf{0.0456 \pm 0.0008}$ & $\mathbf{0.4834 \pm 0.0049}$ & $\mathbf{0.9155 \pm 0.0030}$ & $\mathbf{28.626 \pm 24.219}$ \\
\midrule
A-AVR, W-AVR, MED, FED & \multicolumn{7}{c}{\textit{Fail due to structural dimension mismatch ($K_m \neq K$)}} \\
\bottomrule
\end{tabular}
}
\end{table*}

We generated $100\,000$ data points with ground-truth number of experts $K = 5$, dimension $d = 20$. We simulated $M = 16$ local machines. We evaluated over $50$ independent Monte Carlo replicates. Reported metrics are computed on held-out testing sets. The training--testing split is  $80\%-20\%$.

Three experiments are studied. Experiment~1 is homogeneous, under three mixing-proportion regimes (balanced, moderately imbalanced, and severely imbalanced with $\pi_{\min}=2\%$). Experiment~2 is heterogeneous: each machine observes only $K_m=3$ of the $K=5$ experts. Experiment~3 varies the size $S$ of the supporting sample and its source: covariates pooled across machines, covariates from a single machine, and synthetic covariates drawn from an empirical reference distribution. The data generating process and the settings are described in Appendices~\ref{sec:data_generating_process} and~\ref{Experimental settings}.

\noindent\textbf{Implementation.}
All EM fits use multiple random initializations. The MM algorithm run from two initializations, a warm start from the aligned estimator (A-AVR) and a $k$-means clustering of the pooled local experts, and stops when the decrease of the objective falls below $10^{-6}$ or after $500$ iterations. Details are given in Section~\ref{Experimental settings}.

\noindent\textbf{Results and discussion.}
In the homogeneous benchmark (Table~\ref{tab:official_homogeneous}, Appendix~\ref{Experimental settings}), OT-RED matches the centralized benchmark GLB with more than $100\times$ speedup, and is robust to the label switching that affects W-AVR and to the divergence of FedAvg under subpopulation imbalance ($\pi_{\min}=2\%$). Transportation divergence reaches the centralized benchmark once $S \ge 250$ (Figure~\ref{Figure: sensitivity} and Table~\ref{tab:sensitivity_S}, Appendix~\ref{Experimental settings}).
Under uncoordinated capacities ($K_m=3 \to K=5$, Table~\ref{tab:official_heterogeneous}), all parameter-averaging baselines fail by dimension mismatch, whereas OT-RED aggregates the $48$ local components with accuracy close to the centralized benchmark in $30.7$s, against $4255.7$s for GLB. Its divergence to the pooled density ($\mathcal{T}_c = 104.13$) is the criterion it minimises, and is reported as a diagnostic rather than as a comparison.

\subsection{Real-world data}

\subsubsection{Beijing multi-site air quality}
\label{subsec:real_world_beijing}

We consider the Beijing multi-site air quality dataset \citep{beijing_multi-site_air_quality_501}, where $M=12$ stations record hourly variables over four years ($N=420\,480$,  $d=19$): the data are distributed by construction. OT-RED matches the centralized fit (RMSE $17.76$ against $17.77\,\mu\text{g}/\text{m}^3$) and improves on the averaging baselines, in $17.6$s against $29.6$s; the time gain is modest here because the centralized fit is already fast at this sample size. The task, the out-of-time split and the full table are in Appendix~\ref{appendix: Beijing multi-site air quality}.

\subsubsection{Year prediction MSD}
\label{subsec:real_world_MSD}
We also use the YearPredictionMSD benchmark \citep{BertinMahieux2011MSD} ($N=515\,345$, $d=90$), where the release year of a song is predicted from timbre summaries. Following the official artist-stratified split, the $463\,715$ training tracks are spread over $M=32$ workers, each fitting a MoE with $K=4$ experts, and the aggregation uses $S=2\,000$ support covariates.
\begin{table}[!ht]
\centering
\caption{Performance on the YearPredictionMSD audio benchmark ($M=32$ machines, $K=4$ experts).}
\label{tab:year_msd_benchmark}
\setlength{\tabcolsep}{4pt}
\begin{tabular}{lcccccc}
\toprule
\textbf{Estimator} & $\mathcal{T}_c(\cdot,\bar f^W)$ & \textbf{RMSE (yrs)} & \textbf{MAE (yrs)} & \textbf{RPE} & \textbf{Corr. ($r$)} & \textbf{Time (s)} \\
\midrule
GLB   & $346.4$ & $9.16$ & $6.37$ & $0.7128$ & $0.5365$ & $323.9$ \\
\midrule
OT-RED   & $\mathbf{245.4}$ & $\mathbf{9.27}$ & $\mathbf{6.46}$ & $\mathbf{0.7300}$ & $0.5202$ & $\mathbf{63.5}$ \\
GM    & $307.6$ & $9.34$ & $6.52$ & $0.7412$ & $0.5104$ & $117.8$ \\
A-AVR & $307.1$ & $9.32$ & $6.49$ & $0.7381$ & $\mathbf{0.5207}$ & $117.4$ \\
W-AVR & $311.5$ & $9.46$ & $6.57$ & $0.7604$ & $0.5092$ & $117.4$ \\
MED   & $439.9$ & $9.75$ & $6.62$ & $0.8066$ & $0.4726$ & $117.8$ \\
\bottomrule
\end{tabular}
\end{table}

\noindent\textbf{Results and discussion.}
As shown in Table~\ref{tab:year_msd_benchmark}, OT-RED closely approaches centralized estimation (RMSE $9.27$ vs.\ $9.16$ years, RPE $0.7300$ vs.\ $0.7128$) while being $5.1\times$ faster ($63.5$s vs.\ $323.9$s) and the lowest transportation divergence to the pooled mixture ($\mathcal{T}_c = 245.4$), the criterion it minimises. Details in Appendix~\ref{appendix: Year prediction MSD}.

\section{Conslusion and Discussion}\label{sec: conclusion}

We proposed a one-shot aggregation of distributed mixture-of-experts models: the local models are reduced to a $K$-expert MoE minimizing an expected transportation divergence, by an MM algorithm followed by a fit of the gating network. The estimator is consistent whenever the local estimators are, needs a single communication round, and handles local models with different numbers of experts, although the consistency result assumes $K_m=K$. Choosing $K$ automatically at the server remains open.

\section{Acknowledgement} 
This research was supported by the exploratory research facility ``EXPLO'' at SystemX. 

\renewcommand{\baselinestretch}{1.2}
\normalsize

\section*{References}
\addcontentsline{toc}{section}{References}

\setlength{\bibsep}{6pt}
\bibliographystyle{apalike}
\bibliography{REFERENCES-arxiv}

\newpage
\thispagestyle{empty}
\section*{Appendix}
\addcontentsline{toc}{section}{Appendix}
\addtocontents{toc}{\protect\setcounter{tocdepth}{1}}
\addtocounter{section}{-7}
\renewcommand{\thesection}{\Alph{section}}

This supplementary material provides the proofs of
Propositions \ref{prop: infT infR},
\ref{prop: continuous and convex},
\ref{prop: majorant function},
and Theorem \ref{thm: consistency},
given respectively in Sections
\ref{appendix: proof infT infR},
\ref{appendix: continuous and convex},
\ref{appendix: majorant function},
and \ref{appendix: consistency}.

Sections \ref{apx.update gatingNet} and \ref{apx.update ExpertNet} provide
details on the updating formulas for the gating network and the experts network.
These updates rely on Propositions
\ref{prop: update Gaussian experts} (regression) and
\ref{prop: update logistic experts} (classification),
proved respectively in
Sections \ref{appendix: update Gaussian experts}
and \ref{appendix: update logistic experts}.

Section~\ref{appendix: mass lemma} gives an elementary lemma on transport plans used in several proofs, and Section~\ref{appendix: gating projection} proves Proposition~\ref{prop: gating projection}.

Section \ref{algo.pseudocode} presents the pseudocode of the main algorithm,
and Section \ref{iapx:implementation details and additional experiments}
provides additional implementation details and experimental results.

\section{Proofs}

\subsection{A lemma on transport plans}
\label{appendix: mass lemma}

\begin{lemma}\label{lemma: mass correction}
Let $\Bs p\in\R_+^L$, and let $\Bs a,\Bs b\in\R_+^K$ with $\sum_k a_k=\sum_k b_k=\sum_\ell p_\ell$. Let $\Bs C\in[0,\bar c]^{L\times K}$ and let $\Bs P\in\R_+^{L\times K}$ have row sums $\Bs p$ and column sums $\Bs a$. Then there exists $\Bs Q\in\R_+^{L\times K}$ with row sums $\Bs p$ and column sums $\Bs b$ such that
\[
\sum_{\ell,k}Q_{\ell k}C_{\ell k}\leqslant\sum_{\ell,k}P_{\ell k}C_{\ell k}+\tfrac12\,\bar c\,\|\Bs a-\Bs b\|_1 .
\]
\end{lemma}

\begin{proof}
Let $e_k=(a_k-b_k)^+$, $h_k=(b_k-a_k)^+$ and $\delta=\sum_k e_k=\sum_k h_k=\frac12\|\Bs a-\Bs b\|_1$. If $\delta=0$ take $\Bs Q=\Bs P$. Otherwise, for each column $k$ with $e_k>0$ (hence $a_k>0$) remove the mass $P_{\ell k}e_k/a_k$ from entry $(\ell,k)$; entries stay non-negative and column $k$ now sums to $b_k$. Row $\ell$ has lost $r_\ell=\sum_k P_{\ell k}e_k/a_k$, with $\sum_\ell r_\ell=\delta$. Add $r_\ell h_k/\delta$ to entry $(\ell,k)$ for every $k$. Row sums are back to $\Bs p$ and column $k$ receives $h_k$, so the column sums are $\Bs b$. Removing mass does not increase the cost since $\Bs C\geqslant0$, and the added mass is $\delta$ in total, with unit cost at most $\bar c$.
\end{proof}

\subsection{Proof of Proposition \ref{prop: infT infR}}
\label{appendix: proof infT infR}

In this proof the infima are taken over the class $\widetilde{\mathcal M}_K$ of Proposition~\ref{prop: infT infR}, written $\mathcal M_K$ for short. The second part of the proof sets the gates of $g^\star$ equal to \eqref{eq: app Pi k}, which is always possible in $\widetilde{\mathcal M}_K$ but not with softmax gates; for softmax gates only the first inequality holds, and the gap is controlled by Proposition~\ref{prop: gating projection}.

First, by definition, $\mathcal R_c(g)$ is obtained by relaxing the constraint
in $\Tau_c(g)$ from $\Bs P\in \Pxpipi$ to $\Bs P\in\Pxpi$.
Therefore the inequality
\begin{align*}
\inf_{g\in\mathcal M_K}\Tau_{c}(g)
\geq
\inf_{g\in\mathcal M_K}\mathcal R_c(g)
\end{align*}
holds trivially.

We now prove the reverse inequality.
Let
\[
g^\star
=
\arg\inf_{g\in\mathcal M_K}\mathcal R_c(g)
\]
be a minimizer of $\mathcal R_c(g)$, i.e.,
\[
\mathcal R_c(g^\star)
=
\inf_{g\in\mathcal M_K}\mathcal R_c(g).
\]

We denote the gating and expert functions of $g^\star$ by
$\{\pi^\star_k(\bx), \varphi^\star_k(\cdot|\bx)\}_{k\in[K]}$.

According to \eqref{eq: sum of gating network}, the gating functions satisfy
\begin{align}
\label{eq: app Pi k}
\pi^\star_k(\bx)
=
\sum_{\ell=1}^{L}
\mathcal P_{\ell k}(g^\star, \bx),
\quad \forall \bx\in\mathcal X,
\end{align}
where $\mathcal P_{\ell k}(g^\star, \bx)$ denotes the $(\ell,k)$ entry of
$\mathcal P(g^\star, \bx)$ defined in \eqref{eq: compute optimal plan}.

Equation \eqref{eq: app Pi k} implies that the matrix
$\mathcal P(g^\star, \bx)$ satisfies the marginal constraint with respect to
$\Bs{\pi}^\star(\bx)$, and therefore
\[
\mathcal P(g^\star,\bx)
\in
\Bs{\Pi}_{\bx}(\cdot,\Bs{\pi}^\star).
\]
On the other hand, by definition, $\mathcal P(g^\star,\bx)$ also satisfies the
constraints defining $\Pxpi$.
Consequently,
\[
\mathcal P(g^\star,\bx)
\in
\Bs{\Pi}_{\bx}(\widehat{\Bs{\pi}},\Bs{\pi}^\star)
\quad \forall \bx\in\mathcal X .
\]

Taking expectation yields
\begin{align*}
\inf_{g\in\mathcal M_K}\Tau_{c}(g)
&=
\inf_{g\in\mathcal M_K}
\E\!\left[
\inf_{\Bs P\in\Pxpipi}
\sum_{\ell=1}^{L}\sum_{k=1}^{K}
P_{\ell k}
\,c\!\left(
\widehat\varphi_\ell(\cdot|\bx),
\varphi_k(\cdot|\bx)
\right)
\right]
\\
&\le
\E\!\left[
\sum_{\ell=1}^{L}\sum_{k=1}^{K}
\mathcal P_{\ell k}(g^\star,\bx)
\,c\!\left(
\widehat\varphi_\ell(\cdot|\bx),
\varphi^\star_k(\cdot|\bx)
\right)
\right]
\\
&=
\mathcal R_c(g^\star).
\end{align*}

Combining both inequalities gives
\[
\inf_{g\in\mathcal M_K}\Tau_{c}(g)
=
\inf_{g\in\mathcal M_K}\mathcal R_c(g),
\]
which completes the proof.
\qed

\subsection{Proof of Proposition~\ref{prop: gating projection}}
\label{appendix: gating projection}

Fix $\bx\in\mathcal X$. The plan $\mathcal P(\bar g,\bx)$ has row sums $\widehat{\Bs\pi}(\bx)$ and column sums $\Bs a(\bx)$, and its cost is $\mathcal R_c(\bar g,\bx)$. Lemma~\ref{lemma: mass correction}, with $\Bs b=\Bs\pi(\bx;\bsalpha)$ and $C_{\ell k}=c(\widehat\varphi_\ell(\cdot|\bx),\bar\varphi_k(\cdot|\bx))\leqslant\bar c(\bx)$, gives a plan in $\Bs\Pi_{\bx}(\widehat{\Bs\pi},\Bs\pi(\cdot;\bsalpha))$ with cost at most $\mathcal R_c(\bar g,\bx)+\frac12\bar c(\bx)\|\Bs a(\bx)-\Bs\pi(\bx;\bsalpha)\|_1$. The experts of $\bar f$ are those of $\bar g$, so
\[
\Tau_c(\bar f,\bx)\leqslant\mathcal R_c(\bar g,\bx)+\tfrac12\,\bar c(\bx)\,\|\Bs a(\bx)-\Bs\pi(\bx;\bsalpha)\|_1 .
\]
Taking expectations gives the first inequality. If the experts minimize $\mathcal R_c$, then $\mathcal R_c(\bar g)=\inf_{g}\mathcal R_c(g)=\inf_{g\in\widetilde{\mathcal M}_K}\Tau_c(g)$ by Proposition~\ref{prop: infT infR}, which gives the second.
\qed

\subsection{Barycenter versus reduction}\label{app: barycenter}

With $\bar f^B$, we are finding a $K$-component MoE model $g$ that is the barycenter of the local models $\hf_1,\ldots,\hf_M$ with respect to (w.r.t.)  the weights $\lambda_1,\ldots,\lambda_M$ and the divergence $\rho$.
Whereas with $\bar f^R$, we are finding a $K$-component MoE model $g$ that is closest to the weighted average density $\bar f^W$ defined in \eqref{eq: weighted average f}, w.r.t. the divergence $\rho$.

Because the searching space is $\mathcal M_K$, both $\bar f^B$ and $\bar f^R$ solve the problem of wrong number of components in $\bar f^W$.
 Both are desirable and can be shown to be connected under specific choices of $\rho$.
However, the reduction solution $\bar f^R$ has some advances.
First, as already remarked, the $L$-component MoE $\bar f^W$ is a good approximation to the true conditional density $f^*$, so it is natural to look for a $K$-component MoE model that approximates $\bar f^W$.
Second, the barycenter approach may lead to a counter-intuitive solution under some specific $\rho$, e.g., as already shown in \citep{JMLR-distGMM} for the case of univariate Gaussian mixture models and the $2$-Wasserstein divergence with Euclidean ground distance.
Finally, given the same divergence $\rho$, the computation in \eqref{eq: barycenter approx} will be more expensive than that of \eqref{eq: reduction approx}, especially in our large data context.

\subsection{Assumptions for well-posedness and consistency properties}
\label{appendix: Assumptions}

We make the following standard assumptions before the consistency proposition.
 \begin{enumerate}[label=A\arabic*.]
    \item\label{assumption: iid} The dataset $\mathscr{D} = \{(\bx_i,y_i)\}_{i=1}^N$ is an i.i.d. sample from the $K$-component MoE model $f(y|\bx,\bstheta^*)$ that is 
    ordered and initialized \citep{JIANG19991253}, with distinct expert parameters. 
    \item\label{assumption: continuous} For all $\bx\in\mathcal X$, $c(\varphi_1(\cdot|\bx),\varphi_2(\cdot|\bx))=0$ if and only if $\varphi_1(\cdot|\bx)=\varphi_2(\cdot|\bx)$. This holds for the KL divergence; continuity is assumed in A3.
    \item\label{assumption: convex} The covariate space $\mathcal X$, the expert parameter space $\mathcal B$ (with variances bounded away from zero) and the gating parameter space are compact, and $(\bx,\bsbeta',\bsbeta)\mapsto c(\varphi(\cdot|\bx;\bsbeta'),\varphi(\cdot|\bx;\bsbeta))$ is continuous. In particular $c\leqslant\bar C$ for some constant $\bar C$.
    \item\label{assumption: separation} The distribution of $\bx_0$ has a density on $\R^d$ (the augmented vector $\bx$ lies in the hyperplane $\{x_1=1\}$ and has no density in $\R^{d+1}$), and two experts that coincide on a set of positive $\mu$-measure have the same parameters. This holds for Gaussian regression experts, since two distinct experts coincide at most on a hyperplane.
 \end{enumerate}
Assumption~\ref{assumption: convex} requires bounded covariates. It is not a technicality: with Gaussian experts the KL cost contains the term $(\bx^\top(\bsbeta-\bsbeta'))^2/\sigma^2$, which is unbounded on an unbounded $\mathcal X$, and the uniform bounds used in the proofs fail. The simulated covariates of Section~\ref{section: experimental study} are Gaussian, hence not compactly supported, so the assumption is not met in the strict sense; it can be enforced by truncating at a high quantile, and the covariates of the two real datasets are bounded.
%
%
First, we observe that for any $\bx\in\mathcal X$, the minimization problem inside the expectation operator in \eqref{eq: problem minimize R} is in fact a linear programming (LP) problem. Moreover, for $\bx\in\mathcal X$, we see that $\Pxpi$ is a nonempty set, and the sum
$\sum_{\ell=1}^{L} \sum_{k=1}^K
P_{\ell k}
c\left(\widehat\varphi_\ell(\cdot|\bx),{\varphi}_k(\cdot|\bx)\right)
$
is bounded below by zero for all $\Bs P\in\Pxpi$.
Therefore, this LP problem has a global minimizer (e.g., see \citep{Boydconvex_optimization}). Hence, for all $g\in\mathcal M_K$, there exists a non-negative finite expectation for the random quantity $\mathcal R_{c}(g,{\bx})$.

\subsection{Proof of Proposition \ref{prop: continuous and convex}}
\label{appendix: continuous and convex}

\paragraph{Continuity.}
First we show that, for any fixed $\bx$, the function $\mathcal I(\bsvarphi,\bx)$ is continuous in $g$.
Assumptions \ref{assumption: continuous} and \ref{assumption: convex} imply that the cost function
$c(\cdot,\cdot)$ is continuous in both arguments.
Consequently, the function
\[
\mathcal C(\Bs P,\bsvarphi)
=
\sum_{\ell=1}^{L}\sum_{k=1}^{K}
P_{\ell k}\,
c\!\left(
\widehat\varphi_\ell(\cdot|\bx),
\varphi_k(\cdot|\bx)
\right)
\]
is continuous in $\bsvarphi$ and affine (hence continuous) in $\Bs P$.

Since $\Pxpi$ is a compact set for all $\bx\in\mathcal X$, the function
\[
\bsvarphi \mapsto
\inf_{\Bs P\in\Pxpi}\mathcal C(\Bs P,\bsvarphi)
\]
is continuous, and the infimum is attained.

Moreover, by assumption there exists a constant $\Delta$ such that
$\mathcal I(\bsvarphi,\bx)\le \Delta$ for all $\bsvarphi$ and $\bx$.
Therefore, applying the dominated convergence theorem yields that
for any sequence $\bsvarphi^{(t)}\to \bsvarphi$,
\begin{align*}
\lim_{t\to\infty}
\int_{\mathcal X}
\mathcal I(\bsvarphi^{(t)},\bx)
\,d\mu(\bx)
&=
\int_{\mathcal X}
\lim_{t\to\infty}
\mathcal I(\bsvarphi^{(t)},\bx)
\,d\mu(\bx)
\\
&=
\int_{\mathcal X}
\mathcal I(\bsvarphi,\bx)
\,d\mu(\bx)
=
\mathcal R_c(\bsvarphi).
\end{align*}
Hence $\mathcal R_c(\bsvarphi)$ is continuous.

\paragraph{Bound and existence.}
Since the rows of any $\Bs P\in\Pxpi$ sum to $\widehat\pi_\ell(\bx)$ and $c\leqslant\bar C$ (Assumption~\ref{assumption: convex}), $\mathcal I(\bsvarphi,\bx)\leqslant\bar C\sum_\ell\widehat\pi_\ell(\bx)=\bar C$, so the constant $\Delta$ above can be taken equal to $\bar C$. The infimum over $\Pxpi$ is attained by putting the whole mass of row $\ell$ on a column minimizing $c(\widehat\varphi_\ell(\cdot|\bx),\varphi_k(\cdot|\bx))$, so that
\[
\mathcal I(\bsvarphi,\bx)=\sum_{\ell=1}^{L}\widehat\pi_\ell(\bx)\min_{k\in[K]}c\big(\widehat\varphi_\ell(\cdot|\bx),\varphi_k(\cdot|\bx)\big),
\]
a finite minimum of continuous functions of the expert parameters. $\mathcal R_c$ is therefore continuous on the compact set $\mathcal B^K$ and attains its minimum.

\paragraph{Remark on convexity.} $\mathcal R_c$ is not convex in general: the minimum over $k$ above is not a convex operation, and a convex combination of two elements of $\mathcal M_K$ is a MoE with $2K$ components, which is not in $\mathcal M_K$. Convexity is not needed for existence.

\subsection{Proof of Theorem~\ref{thm: consistency}}
\label{appendix: consistency}

The proof proceeds in three steps.

%
%
%

\paragraph{Notation.}
Because of Assumption~\ref{assumption: iid} and its consistency assumption, every $\widehat{\boldsymbol{\theta}}_m$ has exactly $K$ experts.
Write $\ell=(m,k)$ for the component of $\bar f^W$ coming from expert $k$ of machine $m$. By assumption, for each $m$ there is a permutation $\tau_m$ of $[K]$ such that $\widehat\bsbeta^{(m)}_k\xrightarrow{p}\bsbeta^*_{\tau_m(k)}$ and $\sup_{\bx\in\mathcal X}|\pi_k(\bx;\widehat\bsalpha^{(m)})-\pi^*_{\tau_m(k)}(\bx)|\xrightarrow{p}0$; set $k(\ell)=\tau_m(k)$. The local models need not be aligned: $\mathcal R_c$ and the reduction estimator do not depend on the labels of the local experts. Let $\theta_E=(\bsbeta_k,\sigma_k^2)_{k\in[K]}\in\mathcal B^K$ denote the expert parameters of $g$, where from now on $\bsbeta$ collects all the parameters of an expert (regression coefficients and variance); $\mathcal R_c$ depends on $g$ only through $\theta_E$. By Assumption~\ref{assumption: convex}, the map $(\bx,\bsbeta',\bsbeta)\mapsto c(\varphi(\cdot|\bx;\bsbeta'),\varphi(\cdot|\bx;\bsbeta))$ is uniformly continuous on the compact set $\mathcal X\times\mathcal B^2$; let $\omega$ be its modulus of continuity in $\bsbeta'$, uniform in $(\bx,\bsbeta)$. Define
\[
\delta_N=\max_{\ell\in[L]}\|\widehat\bsbeta_\ell-\bsbeta^*_{k(\ell)}\|,\qquad
\tilde a_k(\bx)=\sum_{\ell:\,k(\ell)=k}\widehat\pi_\ell(\bx)=\sum_{m=1}^M\lambda_m\pi_{\tau_m^{-1}(k)}(\bx;\widehat\bsalpha^{(m)}),\qquad
\eta_N=\sup_{\bx\in\mathcal X}\|\tilde{\Bs a}(\bx)-\Bs\pi^*(\bx)\|_1 ,
\]
where $\Bs\pi^*(\bx)=\Bs\pi(\bx;\bsalpha^*)$. By consistency of the local estimators, $\delta_N\xrightarrow{p}0$ and $\eta_N\xrightarrow{p}0$.

\paragraph{Step 1: $\Tau_c(\bar f^W,f^*)\xrightarrow{p}0$.}
For $\bx\in\mathcal X$, the plan $\widetilde P_{\ell k}(\bx)=\widehat\pi_\ell(\bx)\mathds 1\{k=k(\ell)\}$ has row sums $\widehat{\Bs\pi}(\bx)$ and column sums $\tilde{\Bs a}(\bx)$, which in general differ from $\Bs\pi^*(\bx)$ at finite $N$, so it is not feasible for $\Tau_c(\bar f^W,f^*,\bx)$. Its cost is at most $\omega(\delta_N)$, since $c(\varphi^*_k,\varphi^*_k)=0$. Lemma~\ref{lemma: mass correction} with $\Bs b=\Bs\pi^*(\bx)$ gives a feasible plan with cost at most $\omega(\delta_N)+\frac12\bar C\eta_N$. Hence $0\leqslant\Tau_c(\bar f^W,f^*)\leqslant\omega(\delta_N)+\frac12\bar C\eta_N\xrightarrow{p}0$. This step is not used below; it shows that $\bar f^W$ is close to $f^*$ for the divergence $\Tau_c$.

\paragraph{Step 2: consistency of the experts.}
Define $M_N(\theta_E)=\mathcal R_c(\bar f^W,g)$, its empirical version $M_{N,S}(\theta_E)$ on $\bsX_S$, and $M(\theta_E)=\mathcal R_c(f^*,g)$. By Section~\ref{appendix: continuous and convex},
\begin{align*}
M_N(\theta_E)&=\E\big[I_N(\bx;\theta_E)\big], & I_N(\bx;\theta_E)&=\sum_{\ell}\widehat\pi_\ell(\bx)\min_k c(\widehat\varphi_\ell,\varphi_k),\\
M(\theta_E)&=\E\big[I(\bx;\theta_E)\big], & I(\bx;\theta_E)&=\sum_{k'}\pi^*_{k'}(\bx)\min_k c(\varphi^*_{k'},\varphi_k),
\end{align*}
where the dependence of the densities on $\bx$ is omitted.

\medskip\noindent\textit{(2a) Identification.} $M\geqslant0$ and $M(\theta^*_E)=0$. Let $M(\theta_E)=0$. Since $\pi^*_{k'}(\bx)>0$, for $\mu$-a.e. $\bx$ and every $k'$ there is $k$ with $\varphi_k(\cdot|\bx)=\varphi^*_{k'}(\cdot|\bx)$. As $K$ is finite, for every $k'$ some $k$ satisfies this on a set of positive $\mu$-measure, and Assumption~\ref{assumption: separation} gives equal parameters. Two distinct true experts cannot be matched to the same $k$ (Assumption~\ref{assumption: iid}), so $\theta_E$ is a permutation of $\theta^*_E$. With the ordering convention, $\theta^*_E$ is the unique zero of $M$. By continuity of $M$ (Proposition~\ref{prop: continuous and convex}) and compactness, for all $\epsilon>0$, $\inf\{M(\theta_E):\ \|\theta_E-\theta^*_E\|\geqslant\epsilon\}>0$.

\medskip\noindent\textit{(2b) Uniform convergence.} Grouping the components of $\bar f^W$ by $k(\ell)$,
\[
I(\bx;\theta_E)=\sum_\ell\widehat\pi_\ell(\bx)\min_k c(\varphi^*_{k(\ell)},\varphi_k)+\sum_{k'}\big(\pi^*_{k'}(\bx)-\tilde a_{k'}(\bx)\big)\min_k c(\varphi^*_{k'},\varphi_k).
\]
Using $|\min_k u_k-\min_k v_k|\leqslant\max_k|u_k-v_k|$ and $\sum_\ell\widehat\pi_\ell(\bx)=1$,
\[
\sup_{\bx\in\mathcal X,\,\theta_E\in\mathcal B^K}|I_N(\bx;\theta_E)-I(\bx;\theta_E)|\leqslant\omega(\delta_N)+\bar C\eta_N\xrightarrow{p}0 .
\]
This bound is uniform in $\bx$, so it also holds for the empirical means on $\bsX_S$. Moreover $\{I(\cdot;\theta_E):\theta_E\in\mathcal B^K\}$ is a class of functions bounded by $\bar C$ and continuous in $\theta_E$ on a compact set, hence Glivenko--Cantelli \citep[e.g.,][Example~19.8]{vandervaart1998}: $\sup_{\theta_E}|S^{-1}\sum_{s}I(\bx_s;\theta_E)-M(\theta_E)|\to0$ almost surely as $S\to\infty$. Therefore $\sup_{\theta_E}|M_{N,S}(\theta_E)-M(\theta_E)|\xrightarrow{p}0$.

\medskip\noindent\textit{(2c) Conclusion.} The experts $\bar\theta_E$ returned by the algorithm satisfy $M_{N,S}(\bar\theta_E)\leqslant\inf M_{N,S}+o_P(1)$ by assumption. By (2a), (2b) and Theorem~5.7 of \citet{vandervaart1998}, $\bar\theta_E\xrightarrow{p}\theta^*_E$, that is $\bar\bsbeta^R_k\xrightarrow{p}\bsbeta^*_k$ and $\bar\sigma^{2R}_k\xrightarrow{p}\sigma^{*2}_k$ for all $k$.

\paragraph{Step 3: consistency of the gating parameter.}
Let $a_k(\bx)=\sum_\ell\mathcal P_{\ell k}(\bar g,\bx)$ as in Definition~\ref{def: reduction estimator}, where by \eqref{eq: update transportation plan} the mass of row $\ell$ goes to $\arg\min_{k'}c(\widehat\varphi_\ell(\cdot|\bx),\bar\varphi^R_{k'}(\cdot|\bx))$. By Step 2, Assumption~\ref{assumption: convex} and the triangle inequality on the parameters,
\[
r_N=\max_{\ell,k'}\sup_{\bx}\big|c(\widehat\varphi_\ell(\cdot|\bx),\bar\varphi^R_{k'}(\cdot|\bx))-c(\varphi^*_{k(\ell)}(\cdot|\bx),\varphi^*_{k'}(\cdot|\bx))\big|\xrightarrow{p}0 .
\]
Let $\gamma(\bx)=\min_{k\neq k'}c(\varphi^*_k(\cdot|\bx),\varphi^*_{k'}(\cdot|\bx))$. By Assumptions~\ref{assumption: iid} and~\ref{assumption: separation}, $\gamma(\bx)>0$ for $\mu$-a.e. $\bx$. If $\gamma(\bx)>2r_N$, the value at $k'=k(\ell)$ is below $r_N$ and the other values are above $\gamma(\bx)-r_N>r_N$, so every row goes to its true component and $\Bs a(\bx)=\tilde{\Bs a}(\bx)$. Otherwise $\|\Bs a(\bx)-\tilde{\Bs a}(\bx)\|_1\leqslant2$. Hence
\[
\frac1S\sum_{s=1}^S\|\Bs a(\bx_s)-\Bs\pi^*(\bx_s)\|_1\leqslant\eta_N+\frac2S\sum_{s=1}^S\mathds 1\{\gamma(\bx_s)\leqslant2r_N\}.
\]
For any $t>0$, on the event $\{2r_N<t\}$ the last term is at most $2S^{-1}\sum_s\mathds 1\{\gamma(\bx_s)\leqslant t\}\to2\mu(\gamma\leqslant t)$, and $\mu(\gamma\leqslant t)\to\mu(\gamma=0)=0$ as $t\to0$. So the left-hand side tends to $0$ in probability.

The gating parameter maximizes $L_S(\bsalpha)=S^{-1}\sum_s\sum_k a_k(\bx_s)\log\pi_k(\bx_s;\bsalpha)$. Let $L(\bsalpha)=\E[\sum_k\pi^*_k(\bx)\log\pi_k(\bx;\bsalpha)]$. On the compact gating parameter space and $\mathcal X$, $|\log\pi_k(\bx;\bsalpha)|\leqslant B$ for some $B$, so
\[
\sup_{\bsalpha}|L_S(\bsalpha)-L(\bsalpha)|\leqslant\frac BS\sum_s\|\Bs a(\bx_s)-\Bs\pi^*(\bx_s)\|_1+\sup_{\bsalpha}\Big|\frac1S\sum_s\sum_k\pi^*_k(\bx_s)\log\pi_k(\bx_s;\bsalpha)-L(\bsalpha)\Big|\xrightarrow{p}0,
\]
the second term by the same Glivenko--Cantelli argument. Finally $L(\bsalpha)-L(\bsalpha^*)=-\E[\KL(\Bs\pi^*(\bx)\Vert\Bs\pi(\bx;\bsalpha))]\leqslant0$, with equality only if $\bx^\top(\bsalpha_k-\bsalpha^*_k)=0$ for $\mu$-a.e. $\bx$ and all $k$ (recall $\bsalpha_K=0$), that is $\bsalpha=\bsalpha^*$ since $\mu$ has a density. Theorem~5.7 of \citet{vandervaart1998} gives $\bar\bsalpha^R\xrightarrow{p}\bsalpha^*$.

\paragraph{Conclusion.} Steps 2 and 3 give $\bar\bstheta^R\xrightarrow{p}\bstheta^*$.
\qed

\paragraph*{Misspecified model and number of experts.} If the data are not generated by a $K$-component MoE, the local (quasi-)maximum likelihood estimators converge, under standard conditions \citep{white1982maximum}, to the parameter $\bstheta^\dagger$ of the $K$-component MoE closest to the data distribution in KL divergence. The proof only uses the consistency of the local estimators towards a common limit, and the ordering and separation conditions of Assumptions~\ref{assumption: iid} and~\ref{assumption: separation} for this limit, so the theorem holds with $\bstheta^*$ replaced by $\bstheta^\dagger$ when $\bstheta^\dagger$ is unique up to the ordering. If $K$ is larger than the true number of experts, the parameters are not identifiable and the theorem does not apply; parameter estimation is then slower even without distribution \citep{nguyen2023demystifying}, and we leave this case aside.

\subsection{Proof of Proposition \ref{prop: majorant function}}
\label{appendix: majorant function}

We prove that $\mathcal S_c(\bsvarphi,\bsvarphi^{(t)})$ defined in \eqref{eq: update majorant function}
is a majorant function of $\mathcal R_c(\bsvarphi)$ defined in \eqref{eq: relaxed objective function}
at $\bsvarphi^{(t)}$.
That is, we show that
\[
\mathcal S_c(\bsvarphi,\bsvarphi^{(t)}) \ge \mathcal R_c(\bsvarphi)
\]
for all $\bsvarphi\in\bsPhi^K$, with equality when $\bsvarphi=\bsvarphi^{(t)}$.

First observe that for all $\bsvarphi\in\bsPhi^K$ and $\bx\in\mathcal X$, the definition of
$\mathcal P_{\ell k}(\bsvarphi^{(t)},\bx)$ implies
\[
\sum_{k=1}^{K} \mathcal P_{\ell k}(\bsvarphi^{(t)},\bx)
=
\widehat{\pi}_\ell(\bx),
\quad \forall \ell\in[L].
\]
Therefore the transportation plan
\[
\Bs P =
\big[
\mathcal P_{\ell k}(\bsvarphi^{(t)},\bx)
\big]_{\ell,k}
\]
belongs to the feasible set $\Pxpi$.

By definition of $\mathcal R_c(\bsvarphi)$,
\[
\mathcal R_c(\bsvarphi)
=
\E\!\left[
\inf_{\Bs P\in\Pxpi}
\sum_{\ell=1}^{L}\sum_{k=1}^{K}
P_{\ell k}
\,c\!\left(
\widehat\varphi_\ell(\cdot|\bx),
\varphi_k(\cdot|\bx)
\right)
\right].
\]
Since $\mathcal S_c(\bsvarphi,\bsvarphi^{(t)})$ evaluates the same objective at a feasible
transportation plan $\Bs P=\mathcal P(\bsvarphi^{(t)},\bx)$, we immediately obtain
\[
\mathcal S_c(\bsvarphi,\bsvarphi^{(t)}) \ge \mathcal R_c(\bsvarphi).
\]

It remains to show that equality holds when $\bsvarphi=\bsvarphi^{(t)}$.
In that case we have
\begin{align}
\mathcal R_c(\bsvarphi^{(t)})
&=
\E\!\left[
\inf_{\Bs P\in\Pxpi}
\sum_{\ell=1}^{L}\sum_{k=1}^{K}
P_{\ell k}
\,c\!\left(
\widehat\varphi_\ell(\cdot|\bx),
\varphi^{(t)}_k(\cdot|\bx)
\right)
\right],
\label{eq: app R at gt}
\\
\mathcal S_c(\bsvarphi^{(t)},\bsvarphi^{(t)})
&=
\E\!\left[
\sum_{\ell=1}^{L}\sum_{k=1}^{K}
\mathcal P_{\ell k}(\bsvarphi^{(t)},\bx)
\,c\!\left(
\widehat\varphi_\ell(\cdot|\bx),
\varphi^{(t)}_k(\cdot|\bx)
\right)
\right].
\end{align}

By construction, the transportation plan
\[
\Bs P =
\big[
\mathcal P_{\ell k}(\bsvarphi^{(t)},\bx)
\big]_{\ell,k}
\]
minimizes the inner optimization problem in \eqref{eq: app R at gt}.
Therefore the two expressions coincide, which implies
\[
\mathcal S_c(\bsvarphi^{(t)},\bsvarphi^{(t)})
=
\mathcal R_c(\bsvarphi^{(t)}).
\]

This proves that $\mathcal S_c(\bsvarphi,\bsvarphi^{(t)})$ is a majorant function of
$\mathcal R_c(\bsvarphi)$ at $\bsvarphi^{(t)}$.
\qed

\subsection{Finite stabilization of the MM algorithm}

\begin{proposition}[Finite stabilization of MM]
\label{prop: MM termination}
Let
\[
\mathcal R_{c,S}(\bsvarphi)
=
\frac{1}{S}\sum_{s=1}^S
\sum_{\ell=1}^{L}
\widehat\pi_\ell(\bx_s)
\min_{k\in[K]}
c\!\left(
\widehat\varphi_\ell(\cdot|\bx_s),
\varphi_k(\cdot|\bx_s)
\right)
\]
denote the empirical version of $\mathcal R_c$ on $\bsX_S$.
At iteration $t$, let
$A_t:[L]\times[S]\to[K]$ denote the assignment induced by
\eqref{eq: update transportation plan}, with ties resolved according
to the rule specified above.
Assume that, at every iteration,
\[
\bsvarphi^{(t+1)}
\in
\arg\min_{\bsvarphi\in\bsPhi^K}
\mathcal S_{c,S}(\bsvarphi,\bsvarphi^{(t)}),
\]
where $\mathcal S_{c,S}$ denotes the empirical version of
$\mathcal S_c$.
Then there exists a finite $T$ such that
$A_{t+1}=A_t$ for all $t\geq T$.
Moreover, for every $t\geq T$,
$\bsvarphi^{(t+1)}$ minimizes the surrogate associated with $A_T$, and
$A_T$ is an optimal assignment for $\bsvarphi^{(t+1)}$.

If, in addition, the minimizer of each empirical surrogate is unique,
or a deterministic minimizer is selected as a function of the
assignment, then
$\bsvarphi^{(t+2)}=\bsvarphi^{(t+1)}$ for all $t\geq T$,
so the MM iterations reach a fixed point after finitely many
iterations.
\end{proposition}
The proof is in Section~\ref{appendix: MM termination}; it is the same argument as for the $k$-means algorithm. The limit is a fixed point, not necessarily a global minimizer of $\mathcal R_c$.

For Gaussian experts, the closed-form update of
Proposition~\ref{prop: update Gaussian experts}, together with the
pseudo-inverse and empty-component conventions described in the
supplement, provides a deterministic surrogate minimizer.
For logistic experts, the surrogate is convex and has a unique
minimizer under the nonsingularity condition stated in
Section~\ref{app.update for classification}. Proposition~\ref{prop: MM termination}
concerns exact surrogate minimization; in practice, the Newton
iterations are stopped numerically once the surrogate has converged.
The fixed point need not be a global minimizer of $\mathcal R_{c,S}$.

\subsection{Proof of Proposition~\ref{prop: MM termination}}
\label{appendix: MM termination}

On $\bsX_S$, a plan given by \eqref{eq: update transportation plan} is described by an assignment $A:[L]\times[S]\to[K]$, and there are finitely many assignments. Write $A_t$ for the assignment of $\bsvarphi^{(t)}$ and, for an assignment $A$, $V(A)=\min_g\mathcal S_c(\bsvarphi,A)$, where $\mathcal S_c(\bsvarphi,A)$ is the empirical majorant computed with the plan of $A$. By construction $\bsvarphi^{(t+1)}$ minimizes $\mathcal S_c(\cdot,A_t)$, so $V(A_t)=\mathcal S_c(\bsvarphi^{(t+1)},A_t)$, and
\[
V(A_{t+1})\leqslant\mathcal S_c(\bsvarphi^{(t+1)},A_{t+1})=\mathcal R_c(\bsvarphi^{(t+1)})\leqslant\mathcal S_c(\bsvarphi^{(t+1)},A_t)=V(A_t).
\]
If $A_{t+1}\neq A_t$, the tie-breaking rule implies that for some $(\ell,s)$ the new expert has a strictly smaller cost than the previous one, and since $\widehat\pi_\ell(\bx_s)>0$ the second inequality is strict, so $V(A_{t+1})<V(A_t)$. Hence no assignment can appear twice unless the sequence has become constant, and since there are finitely many assignments, $A_{t+1}=A_t$ after a finite number of iterations. From then on $\bsvarphi^{(t+1)}$ minimizes $\mathcal S_c(\cdot,A_t)$ and $A_t$ is its own optimal assignment.
\qed

\section{Details of updating the gating and expert networks of the aggregated MoE model}
\label{app: DETAILS OF UPDATING}

\subsection{Updating the gating network of the distributed MoE model}
\label{apx.update gatingNet}

Up to this point, the optimizations performed on the central machine involve expectations with respect to $\bx$. In practice, the distribution of $\bx$ is unknown.
Therefore, we approximate these expectations using a supporting sample
$\bsX_S=\{(\bx_s,y_s)\}_{s=1}^S$ obtained by randomly drawing observations from the full dataset $\sD$.
This sample $\bsX_S$ is used to approximate the expectation operator on the central machine.

Specifically, the expectation $\E_{\mu(\bx)}[\cdot]$ is approximated by the empirical expectation based on the empirical distribution
\[
\mu(\cdot)=\frac{1}{S}\sum_{s=1}^S\delta_{\bx_s}(\cdot).
\]

The choice of the size $S$ is discussed in the experimental section.
In the following, we assume that $S$ is sufficiently large to approximate the distribution of $\bx$.

We denote by $\Bs X_S$ the set of covariate observations $\bx_s$ in the supporting sample $\bsX_S$.
When no confusion arises, $\Bs X_S$ also denotes the matrix whose rows correspond to the row vectors $\bx_s^\top$.

Although equation \eqref{eq: sum of gating network} provides the mixing weights for any covariate $\bx$, obtaining the final MoE model requires estimating the parameter $\bar{\bsalpha}^R$ of the gating functions $\pi_k(\bx;\bar{\bsalpha}^R)$.
We estimate $\bar{\bsalpha}^R$ by solving a softmax regression problem in which the predictors are $\Bs X_S$ and the responses are computed according to \eqref{eq: sum of gating network}.

Specifically, let $\Bs A=[a_{sk}]$ be a matrix in $\R^{S\times K}$ defined by
\begin{equation*}
\label{eq: response a_sk}
a_{sk} = \sum_{\ell=1}^{L}\mathcal P_{\ell k}(\bar g,\bx_s),
\end{equation*}
where $\bar g$ denotes the model obtained at convergence of the MM algorithm, $\bx_s\in\bsX_S$, and $\mathcal P_{\ell k}(\bar g,\bx_s)$ is computed according to \eqref{eq: update transportation plan}.

The gating parameter $\bar{\bsalpha}^R$ is then estimated via maximum likelihood as
\begin{align}
\label{eq: calculate gating network parameter}
\bar{\bsalpha}^R
&= \arg\max_{\bsalpha}  \sum_{s=1}^S \sum_{k=1}^{K} a_{sk}\log \pi_{k}(\bx_s;\bsalpha) \nonumber\\
&= \arg\max_{\bsalpha}  \sum_{s=1}^S
\left[
\sum_{k=1}^{K-1} a_{sk}\bsalpha^\top_{k}\bx_s
- \log\left(1 + \sum_{k^\prime =1}^{K-1}  \exp\{\bsalpha^\top_{k^\prime}\bx_s\} \right)
\right].
\end{align}

This optimization problem can be solved efficiently using, for example, the Newton--Raphson procedure.
The parameter vector is initialized by fixing the $K$-th gate parameter to zero, which is the standard identifiability constraint for softmax regression and is required for establishing the consistency of $\bar \bstheta^{R}$.

We now turn to the problem of updating the expert parameters $\bsbeta^{(t+1)}_k$ defined in \eqref{eq: update expert k}.
In the following subsections we derive the update formulas for two common specifications of experts: Gaussian regression experts and logistic regression experts when the cost function $c(\cdot,\cdot)$ is the KL-divergence.

\subsection{Updating the Gaussian regression expert parameters}
\label{apx.update ExpertNet}

We now consider the case where the experts are Gaussian regression models, i.e.
\[
\varphi(y|\bx) = \varphi(y|\bx^\top\bsbeta; \sigma^2),
\]
where $\bsbeta\in\R^{d+1}$ and $\sigma^2\in\R_+$ denote the parameters of the expert.

The KL-divergence between two Gaussian regression experts takes the following form
\begin{equation*}
\label{eq: conditional KL Gaussian}
\KL\big(\varphi_1(\cdot|\bx)\Vert \varphi_2(\cdot|\bx)\big)
=
\frac{1}{2}
\left(
\log \frac{\sigma_2^2}{\sigma_1^2}
+
\frac{\sigma_1^2}{\sigma_2^2}
+
\frac{(\bsbeta_2-\bsbeta_1)^\top \bx\bx^\top(\bsbeta_2-\bsbeta_1)}{\sigma_2^2}
-1
\right).
\end{equation*}

Therefore, when the experts are Gaussian and the cost function $c(\cdot,\cdot)$ is chosen as the KL-divergence, the objective function in problem \eqref{eq: update expert k} can be written as a function of $\bsbeta$ and $\sigma^2$:
\begin{align}
\label{eq: K function Gaussian}
\mathcal K(\bsbeta,\sigma^2)
\colonequals
\E\!\left[
\sum_{\ell=1}^{L}
\mathcal P^{(t)}_{\ell k}(\bx)
\frac{1}{2}
\left(
\log \frac{\sigma^2}{\widehat\sigma_\ell^2}
+
\frac{\widehat\sigma_\ell^2}{\sigma^2}
+
\frac{(\bsbeta-\widehat\bsbeta_\ell)^\top \bx\bx^\top(\bsbeta-\widehat\bsbeta_\ell)}{\sigma^2}
-1
\right)
\right],
\end{align}
where $\mathcal P^{(t)}_{\ell k}(\bx)$ denotes $\mathcal P_{\ell k}(g^{(t)},\bx)$ computed according to \eqref{eq: update transportation plan} with the cost replaced by the KL-divergence.

Hence, problem \eqref{eq: update expert k} becomes
\[
(\bsbeta_k^{(t+1)},\sigma_k^{2(t+1)})
=
\arg\min_{(\bsbeta,\sigma^2)\in\R^{d+1}\times\R_+}
\mathcal K(\bsbeta,\sigma^2).
\]

The following proposition provides the update formulas for the parameters of the Gaussian regression experts.

\begin{proposition}
\label{prop: update Gaussian experts}
Let the notations be as defined above. If $c(\cdot,\cdot)$ is chosen as $\KL(\cdot\Vert\cdot)$ and $\varphi_k^{(t+1)}(\cdot|\bx)$ in \eqref{eq: update expert k} are Gaussian regression experts, then the parameter updates are given by
\begin{subequations}
\label{eq: update Gaussian experts}
{\small \begin{align}
\bsbeta_k^{(t+1)}
&=
(\Bs X_S^\top \Bs D_k^{(t)} \Bs X_S)^{-1}
\Bs X_S^\top \Bs r_k^{(t)},
\qquad \Bs r_k^{(t)}=\big((\Bs X_S\widehat{\Bs B})\odot\Bs W_k^{(t)\top}\big)\mathds 1_{L},
\\
\sigma_k^{2(t+1)}
&=
\frac{1}{\trace(\Bs D_k^{(t)})}
\left(
\widehat{\Bs\Sigma}^\top \Bs W_k^{(t)}\mathds{1}_S
+
\mathds{1}_{L}^\top
\big(\Bs W_k^{(t)}\odot\Bs E_k\odot\Bs E_k\big)
\mathds{1}_{S}
\right),
\quad \Bs E_k=\big(\bsbeta_k^{(t+1)}\mathds 1_{L}^\top-\widehat{\Bs B}\big)^\top\Bs X_S^\top.
\end{align}}
\end{subequations}Here the matrices $\widehat{\Bs B}$, $\Bs D_k^{(t)}$, and $\Bs W_k^{(t)}$  are defined in \eqref{eq: matrix B}, \eqref{eq: matrix D}, and \eqref{eq: matrix W}, respectively.
\end{proposition}

The detailed derivations are provided in Section~\ref{appendix: update Gaussian experts}.
If $\Bs X_S^\top\Bs D_k^{(t)}\Bs X_S$ is singular, the inverse is replaced by the Moore--Penrose pseudo-inverse, which still gives a minimizer; if expert $k$ receives no mass, its contribution to $\mathcal S_c$ is zero and it is left unchanged. In both cases the update minimizes $\mathcal S_c$, so the descent property and Proposition~\ref{prop: MM termination} hold.

\subsection{Proof of Proposition \ref{prop: update Gaussian experts}}
\label{appendix: update Gaussian experts}

To derive the update formulas, we minimize the function $\mathcal K(\bsbeta,\sigma^2)$ defined in \eqref{eq: K function Gaussian}.

First, we observe that $\mathcal K(\bsbeta,\sigma^2)$ is a quadratic convex function with respect to $\bsbeta$, and therefore any stationary point is a global minimizer with respect to $\bsbeta$.
Moreover, $\mathcal K(\bsbeta,\sigma^2)$ is coercive with respect to $\sigma^2$, which guarantees the existence of a minimizer satisfying the first-order optimality conditions.

The partial derivatives of $\mathcal K(\bsbeta,\sigma^2)$ with respect to $\bsbeta$ and $\sigma^2$ are
\begin{align*}
\frac{\partial  \mathcal K}{\partial \bsbeta}
&= \E
\left[
\sum_{\ell=1}^{L}
\mathcal P_{\ell k}^{(t)}(\bx)
\frac{\bx\bx^\top (\bsbeta - \widehat\bsbeta_\ell)}{\sigma^2}
\right],
\\
\frac{\partial  \mathcal K}{\partial \sigma^2}
&= \E
\left[
\sum_{\ell=1}^{L}
\mathcal P_{\ell k}^{(t)}(\bx)
\left(
\frac{\sigma^2 - \widehat\sigma_\ell^2 - (\bsbeta - \widehat\bsbeta_{\ell})^\top \bx \bx^\top (\bsbeta - \widehat\bsbeta_{\ell})}
{2\sigma^4}
\right)
\right].
\end{align*}

Replacing the expectation by the empirical expectation over the supporting sample and setting $\frac{\partial  \mathcal K}{\partial \bsbeta}=0$ gives
\begin{align*}
\sum_{s=1}^S
\bx_s
\left(
\sum_{\ell=1}^{L}
\mathcal P_{\ell k}^{(t)}(\bx_s)
\right)
\bx_s^\top
\bsbeta
=
\sum_{s=1}^S
\bx_s\bx_s^\top
\sum_{\ell=1}^{L}
\mathcal P_{\ell k}^{(t)}(\bx_s)
\widehat \bsbeta_{\ell}.
\end{align*}

Let
\begin{align}
\widehat{\Bs B} &=
\left[ \widehat\bsbeta_1, \ldots, \widehat\bsbeta_{L} \right]
\in\R^{p\times L},
\label{eq: matrix B}
\\
\Bs D_k^{(t)} &=
\diag\left(
\sum_{\ell=1}^{L}\mathcal P_{\ell k}^{(t)}(\bx_1),
\ldots,
\sum_{\ell=1}^{L}\mathcal P_{\ell k}^{(t)}(\bx_S)
\right)
\in[0,1]^{S\times S},
\label{eq: matrix D}
\\
\Bs W_k^{(t)} &=
\begin{bmatrix}
\mathcal P_{1k}^{(t)}(\bx_1) & \cdots & \mathcal P_{1k}^{(t)}(\bx_S)\\
\vdots & \ddots & \vdots\\
\mathcal P_{L,k}^{(t)}(\bx_1) & \cdots & \mathcal P_{L,k}^{(t)}(\bx_S)
\end{bmatrix}
\in[0,1]^{L\times S}.
\label{eq: matrix W}
\end{align}

The right-hand side is $\sum_s\bx_s r_s$ with $r_s=\sum_\ell\mathcal P^{(t)}_{\ell k}(\bx_s)\,\bx_s^\top\widehat\bsbeta_\ell$, the $s$-th entry of $\Bs r_k^{(t)}=((\Bs X_S\widehat{\Bs B})\odot\Bs W_k^{(t)\top})\mathds 1_{L}$, since $(\Bs X_S\widehat{\Bs B})_{s\ell}=\bx_s^\top\widehat\bsbeta_\ell$ and $(\Bs W_k^{(t)\top})_{s\ell}=\mathcal P^{(t)}_{\ell k}(\bx_s)$. The previous equation is therefore
\[
\Bs X_S^\top \Bs D_k^{(t)} \Bs X_S \bsbeta=\Bs X_S^\top\Bs r_k^{(t)},
\]
and, when $\Bs X_S^\top \Bs D_k^{(t)} \Bs X_S$ is invertible,
$\bsbeta=(\Bs X_S^\top \Bs D_k^{(t)} \Bs X_S)^{-1}\Bs X_S^\top\Bs r_k^{(t)}$.

Similarly, setting $\frac{\partial\mathcal K}{\partial\sigma^2}=0$ and replacing the expectation by the empirical mean gives
\[
\sigma^2\sum_{s,\ell}\mathcal P^{(t)}_{\ell k}(\bx_s)=\sum_{s,\ell}\mathcal P^{(t)}_{\ell k}(\bx_s)\Big(\widehat\sigma_\ell^2+\big(\bx_s^\top(\bsbeta-\widehat\bsbeta_\ell)\big)^2\Big).
\]
The left-hand side is $\trace(\Bs D_k^{(t)})\sigma^2$. On the right, the first sum is $\widehat{\Bs\Sigma}^\top\Bs W_k^{(t)}\mathds 1_S$ with $\widehat{\Bs\Sigma}=(\widehat\sigma_1^2,\ldots,\widehat\sigma_{L}^2)^\top$, and the second is $\mathds 1_{L}^\top(\Bs W_k^{(t)}\odot\Bs E_k\odot\Bs E_k)\mathds 1_S$ with $(\Bs E_k)_{\ell s}=\bx_s^\top(\bsbeta-\widehat\bsbeta_\ell)$.

This yields the update formula for $\sigma^2$ and completes the proof.
\qed

\subsection{Updating the logistic regression expert parameters}
\label{app.update for classification}

When the task is classification, it is common to model the expert components using logistic regression models. For simplicity, we consider the case of binary responses, i.e. $y\in\{0,1\}$. The conditional density then takes the form
\begin{equation*}
\varphi(y|\bx;\bsbeta)
=
\left[\frac{\exp(\bx^\top \bsbeta)}{1+\exp(\bx^\top \bsbeta)}\right]^y
\left[\frac{1}{1+\exp(\bx^\top \bsbeta)}\right]^{1-y},
\end{equation*}
where $\bsbeta\in\R^{d+1}$ is the parameter vector.

Let $\varphi_1(y|\bx;\bsbeta_1)$ and $\varphi_2(y|\bx;\bsbeta_2)$ denote the conditional densities of two binary logistic regression experts. The conditional KL-divergence between them is
\begin{align}
\label{eq: conditional KL logistic}
\KL\big(\varphi_1(\cdot|\bx)\Vert \varphi_2(\cdot|\bx)\big)
&=
\frac{\exp(\bx^\top \bsbeta_1)}{1+\exp(\bx^\top \bsbeta_1)}
\left(
\log \frac{\exp(\bx^\top \bsbeta_1)}{1+\exp(\bx^\top \bsbeta_1)}
-
\log\frac{\exp(\bx^\top \bsbeta_2)}{1+\exp(\bx^\top \bsbeta_2)}
\right)
\nonumber\\
&\quad+
\frac{1}{1+\exp(\bx^\top \bsbeta_1)}
\left(
\log \frac{1}{1+\exp(\bx^\top \bsbeta_1)}
-
\log \frac{1}{1+\exp(\bx^\top \bsbeta_2)}
\right).
\end{align}

Similarly to the Gaussian expert case, substituting the KL-divergence expression into \eqref{eq: update expert k} leads to the following objective function with respect to $\bsbeta$:
\begin{align}
\label{eq: J function logistic}
\mathcal K_{\text{log}}(\bsbeta)
=
\E\left[
\sum_{\ell=1}^{L}
\mathcal P^{(t)}_{\ell k}(\bx)
\left(
\log\!\left(1+\exp(\bx^\top \bsbeta)\right)
-
\bx^\top\bsbeta
\frac{\exp(\bx^\top \widehat\bsbeta_\ell)}
{1+\exp(\bx^\top \widehat\bsbeta_\ell)}
+ \text{const}
\right)
\right],
\end{align}
where $\mathcal P^{(t)}_{\ell k}(\bx)$ is computed as in \eqref{eq: update transportation plan}, with the cost function replaced by the KL-divergence \eqref{eq: conditional KL logistic}.

\begin{proposition}
\label{prop: update logistic experts}
Let $d_s=\sum_{\ell=1}^{L}\mathcal P^{(t)}_{\ell k}(\bx_s)$ (the diagonal of $\Bs D_k^{(t)}$), and for $d_s>0$ let
\[
v_s=\frac1{d_s}\sum_{\ell=1}^{L}\mathcal P^{(t)}_{\ell k}(\bx_s)\,\frac{\exp(\bx_s^\top\widehat\bsbeta_\ell)}{1+\exp(\bx_s^\top\widehat\bsbeta_\ell)}\in(0,1),
\qquad\text{i.e. } \Bs v_k^{(t)}=\Bs D_k^{(t)-1}\big(\Bs W_k^{(t)\top}\odot\Bs U\big)\mathds 1_{L},
\]
where $\Bs U\in(0,1)^{S\times L}$ has entries $u_{s\ell}=\exp(\bx_s^\top\widehat\bsbeta_\ell)/(1+\exp(\bx_s^\top\widehat\bsbeta_\ell))$. If $c(\cdot,\cdot)$ is the KL divergence and the experts are logistic regressions, then, with the expectation replaced by the empirical mean on $\bsX_S$, problem \eqref{eq: update expert k} is the weighted logistic regression with fractional responses
\begin{equation}\label{eq: update logistic experts}
\bsbeta_k^{(t+1)}\in\arg\min_{\bsbeta}\ \sum_{s=1}^S d_s\Big(\log\big(1+\exp(\bx_s^\top\bsbeta)\big)-v_s\,\bx_s^\top\bsbeta\Big).
\end{equation}
The objective is convex, strictly convex when $\Bs X_S^\top\Bs D_k^{(t)}\Bs X_S$ is invertible, and has no closed-form minimizer in general. It is minimized by Newton (IRLS) iterations
\begin{equation}\label{update beta logistic}
\bsbeta\leftarrow\bsbeta+\big(\Bs X_S^\top\Bs D_k^{(t)}\Bs\Lambda(\bsbeta)\Bs X_S\big)^{-1}\Bs X_S^\top\Bs D_k^{(t)}\big(\Bs v_k^{(t)}-\psi(\Bs X_S\bsbeta)\big),
\end{equation}
where $\psi(u)=e^u/(1+e^u)$ is the logistic function applied entrywise and $\Bs\Lambda(\bsbeta)=\diag\big(\psi(\bx_s^\top\bsbeta)(1-\psi(\bx_s^\top\bsbeta))\big)$, started at $\bsbeta_k^{(t)}$.
\end{proposition}

The MM descent property only requires that the update does not increase the majorant $\mathcal S_c(\bsvarphi,\bsvarphi^{(t)})$, so a few Newton steps with a step-size check are enough in practice.

\subsection{Proof of Proposition \ref{prop: update logistic experts}}
\label{appendix: update logistic experts}

Replacing the expectation in \eqref{eq: J function logistic} by the empirical mean on $\bsX_S$ and dropping the constant, the objective is
\[
\sum_{s=1}^S\sum_{\ell=1}^{L}\mathcal P^{(t)}_{\ell k}(\bx_s)\Big(\log\big(1+\exp(\bx_s^\top\bsbeta)\big)-\bx_s^\top\bsbeta\,u_{s\ell}\Big)
=\sum_{s=1}^S d_s\Big(\log\big(1+\exp(\bx_s^\top\bsbeta)\big)-v_s\,\bx_s^\top\bsbeta\Big),
\]
where $u_{s\ell}$ is the $(s,\ell)$ entry of $\Bs U$; rows with $d_s=0$ do not contribute. The gradient is $\Bs X_S^\top\Bs D_k^{(t)}(\psi(\Bs X_S\bsbeta)-\Bs v_k^{(t)})$ and the Hessian is $\Bs X_S^\top\Bs D_k^{(t)}\Bs\Lambda(\bsbeta)\Bs X_S$, which is positive semi-definite, and positive definite when $\Bs X_S^\top\Bs D_k^{(t)}\Bs X_S$ is invertible because the diagonal of $\Bs\Lambda(\bsbeta)$ is positive. The first-order condition $\Bs X_S^\top\Bs D_k^{(t)}\psi(\Bs X_S\bsbeta)=\Bs X_S^\top\Bs D_k^{(t)}\Bs v_k^{(t)}$ is a system of $d+1$ nonlinear equations in $\bsbeta$; since $\Bs X_S^\top$ has more columns than rows, it does not imply $\psi(\Bs X_S\bsbeta)=\Bs v_k^{(t)}$, and there is no closed form in general. Since $v_s\in(0,1)$, the objective tends to $+\infty$ in every direction $\bsbeta=\rho\Bs u$, $\rho\to\infty$, with $\Bs D_k^{(t)1/2}\Bs X_S\Bs u\neq0$, so a minimizer exists when $\Bs X_S^\top\Bs D_k^{(t)}\Bs X_S$ is invertible. The Newton step above is the usual IRLS step for this objective.
\qed

\section{Pseudocode of the aggregated algorithm}\label{algo.pseudocode}

Algorithm~\ref{algo: main algorithm} summarizes the aggregation procedure to obtain the proposed OT-based reduction estimator.

\begin{algorithm}[!h]
\caption{One-shot optimal transport aggregation for distributed MoE}
\label{algo: main algorithm}
\begin{algorithmic}[1]

\STATEx{
\textbf{Input:}
\begin{itemize}[itemsep=-2pt]
\item local estimators $\{\widehat{\bstheta}_m\}_{m=1}^M$ with $\widehat{\bstheta}_m = (\widehat{\bsalpha}^{(m)},\widehat{\bsbeta}_1^{(m)}, \ldots, \widehat{\bsbeta}_{K_m}^{(m)})$;
\vspace{2.5pt}
\item sample proportions $\lambda_m = N_m / N$, $m\in[M]$;
\vspace{2.5pt}
\item supporting sample $\bsX_S = \{\bx_s\}_{s=1}^S$, e.g., by drawing from $\sD$  (if the real-world problem setting allows to do that); or by simulating using the parameters of some parametric distributions fitted on covariates data on the local machines;
\item $K$, the number of experts of the target aggregated MoE model.
\end{itemize}
}

\STATEx{
\textbf{Output:}
Aggregated reduction estimator
\(
\bar\bstheta^{R} =
(\bar{\bsalpha}^R, \bar{\bsbeta}_1^R, \ldots, \bar{\bsbeta}_K^R).
\)
}

\vspace{0.2cm}\hrule\vspace{0.15cm}

\STATE \textbf{Initialize}: Set $t \leftarrow 0$. Initialize target expert parameters
${\bsbeta}_1^{(0)}, \ldots, {\bsbeta}_K^{(0)}$, e.g., via aligned
averaging or $K$-means clustering on local expert parameter vectors.
\STATE \textbf{STAGE 1:  Expert reduction via Majorization–Minimization}







\REPEAT
    \STATE (\textit{Find optimal transport plan})
    \FORALL{$\ell \in [L]$ and $s \in [S]$}
        \STATE Find best matching expert: $k^*(\ell, s) \leftarrow \arg\min_{k \in [K]} c\big(\widehat{\varphi}_\ell(\cdot \mid \bx_s),\, \varphi_k^{(t)}(\cdot \mid \bx_s)\big)$.
        \STATE Assign mass (see \eqref{eq: update transportation plan}): $\mathcal{P}_{\ell k}^{(t)}(\bx_s) \leftarrow \widehat{\pi}_\ell(\bx_s) \cdot \mathbb{I}_{\{k = k^*(\ell, s)\}}$.
    \ENDFOR

    \STATE (\textit{Update expert parameters})
    \FOR{$k\in[K]$}
        \STATE
\parbox[t]{\dimexpr0.97\linewidth-\algorithmicindent\relax}{%
Update the expert parameter
${\bsbeta}_k^{(t+1)}$ by solving \eqref{eq: update expert k}.
For Gaussian expert, the closed-form formula for updating
${\bsbeta}_k^{(t+1)}$ is given in
\eqref{eq: update Gaussian experts}. For multinomial logistic regression expert, ${\bsbeta}_k^{(t+1)}$ is
obtained by solving a weighted logistic regression problem defined
in \eqref{eq: update logistic experts}, whose updating formula for
the IRLS iteration is given in \eqref{update beta logistic}.
}
    \ENDFOR
    \STATE $t \leftarrow t + 1$.
\UNTIL{$\big|\mathcal{R}_{c,S}(\bsvarphi^{(t)}) - \mathcal{R}_{c,S}(\bsvarphi^{(t-1)})\big| \leqslant \epsilon_{\mathrm{tol}}$ \textbf{or} plan $\mathcal{P}^{(t)}$ stabilizes (Proposition~\ref{prop: MM termination}).}

\STATE Set the converged experts: $\bar{\bsbeta}_k^R \leftarrow \bsbeta_k^{(t^*)}$ for all $k \in [K]$.
\vspace{0.08cm}

\STATE \textbf{STAGE 2: Gating projection via Newton–Raphson}
\STATE Compute empirical target gate marginals $\mathbf{a}_s$ on $\bsX_S$:
$
a_{sk} \leftarrow \sum_{\ell=1}^L \mathcal{P}_{\ell k}^{(t)}(\bx_s)$, for all $s \in [S], \; k \in [K].
$
\vspace{0.05cm}
\STATE Estimate $\bar{\bsalpha}^R$ by solving the multinomial logistic regression \eqref{eq: calculate gating network parameter} via Newton--Raphson using $\{(\bx_s, \mathbf{a}_s)\}_{s=1}^S$.
\vspace{0.1cm}
\end{algorithmic}
\end{algorithm}

\normalsize

\section{Experimental details and additional results}
\label{iapx:implementation details and additional experiments}
This appendix comprises the material supporting Section~\ref{section: experimental study}.
Section~\ref{Example} illustrates, on a one-dimensional example, what the reduction
estimator recovers and where the weighted average fails.
Section~\ref{Baselines and metrics} defines the baselines and the evaluation metrics,
Section~\ref{sec:data_generating_process} the simulation design, and
Section~\ref{Experimental settings} reports the detailed simulation results, including the
homogeneous benchmark and the sensitivity to the supporting sample.
Section~\ref{appendix: Beijing multi-site air quality} and
Section~\ref{appendix: Year prediction MSD} describe the two real datasets, their splits
and the corresponding results.

\subsection{Example of reduction solution}\label{Example}
We first illustrate the aggregation problem on a one-dimensional example with $M=2$
machines and $K=2$ Gaussian experts with a softmax gate. Each machine fits its own MoE on
its own sample, and the two local fits order their experts differently: expert~$1$ of the
first machine plays the role of expert~$2$ of the second machine. This is the label
switching situation described in Section~\ref{sec: core technical}, and it is the rule
rather than the exception, since nothing ties the labels of two models estimated
independently.

The top row of Figure~\ref{Figure: toy illustration} shows the regression functions. The
weighted average $\bar f^W$ carries the $L=MK=4$ local experts (left), so the structure of
the model is lost even though the resulting conditional density still approximates $f^*$
reasonably well: as a density it is accurate, as an estimator of $\bstheta^*$ it is not.
The reduction $\bar f^R$ (right) returns $K=2$ experts, which coincide with the two true
expert lines.

The bottom row shows the gating functions, where the difference is sharper. Averaging the
local gates mixes gates that refer to different experts, and the resulting weights
$\sum_m \lambda_m \widehat\pi^{(m)}_k$ are almost constant and close to $1/2$ over the whole
covariate range (left): the averaged model has lost the covariate-dependent assignment that
defines the MoE. The gates recovered from the OT plan (right) follow the true
gates $\pi^*_1$ and $\pi^*_2$, because the plan matches local experts by their conditional
densities and not by their index.

\begin{figure}[!h]
\centering
\subfloat{\includegraphics[width=.36\linewidth]{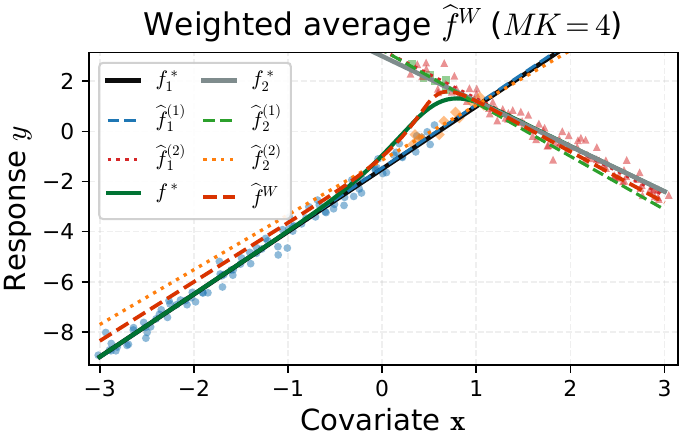}}\hspace{.5cm}
\subfloat{\includegraphics[width=.36\linewidth]{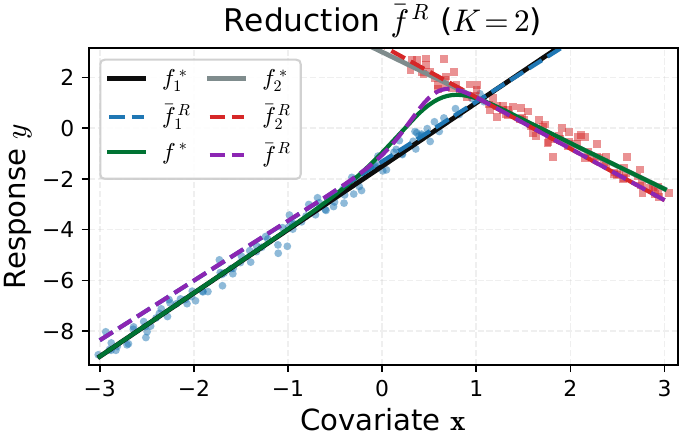}}\\
\subfloat{\includegraphics[width=.36\linewidth]{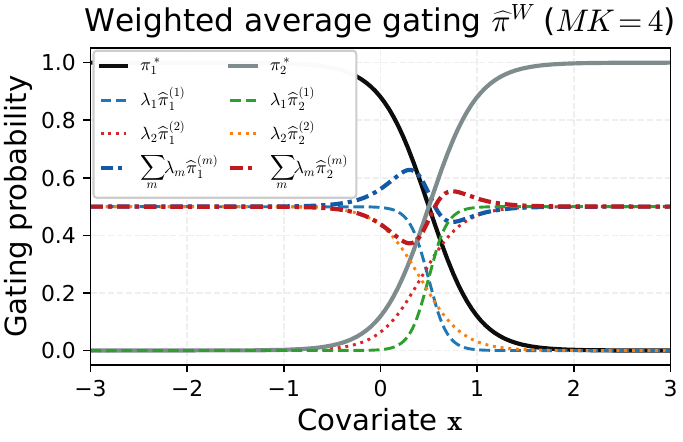}}\hspace{.5cm}
\subfloat{\includegraphics[width=.36\linewidth]{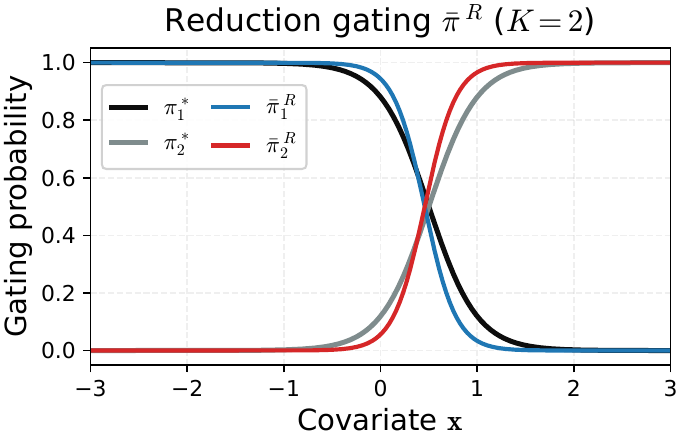}}\\
\caption{One-dimensional example with $M=2$ machines and $K=2$ experts. Top: regression
functions; bottom: gating functions. Left: the weighted average $\bar f^W$ has $L=MK=4$
experts and fails to recover the gates, which collapse to $1/2$, due to label switching.
Right: the reduction $\bar f^R$ (see Section~\ref{proposed aggregation strategy}) has $K=2$
experts, with gates recovered from the transport plan, and is robust to local label
switching.}

\label{Figure: toy illustration}
\end{figure}

\subsection{Baselines and metrics}\label{Baselines and metrics}

Interpretations of the baselines presented in the experimental studies:
\begin{itemize}[label=--]
\item Proposed reduction $\bar\bstheta^R$ (OT-RED): Our one-shot aggregation minimizing expected conditional transportation divergence $\mathbb E_{\bx}[\Tau_c(\bar f^W(\cdot|\bx), f(\cdot|\bx;\bstheta))]$ on supporting sample $\bsX_S$. Expert parameters $(\beta_k, \sigma_k^2)_{k=1}^K$ are fitted via Majorization-Minimization on $\mathcal R_c$, and softmax gates $(\alpha_k)_{k=1}^K$ via IRLS on the transport plan marginals.

\item Centralized benchmark $\bstheta^G$ (GLB): Centralized MLE $\arg\max_{\bstheta} \sum_{i=1}^N \log f(y_i|\bx_i;\bstheta)$ trained on pooled data $\bigcup_{m=1}^M \sD_m$ via EM, serving as the theoretical upper bound.

\item Greedy merging $\bar\bstheta^{GM}$ (GM): Iteratively merges the pair of components in $\bar f^W$ minimizing pairwise conditional KL divergence cost on $\bsX_S$ until down to $K$ experts \citep{runnalls2007kullback}, the gates are then fitted as for OT-RED.

\item Aligned average $\bar\bstheta^A$ (A-AVR): The experts of each local model are first matched to those of machine 1 (reference) using Hungarian algorithm on the KL divergence between experts, the matched parameters are then averaged. This removes the label switching problem of $\bar\bstheta^{W}$ and strictly requires $K_m = K$.

\item Federated average $\bar\bstheta^{\mathrm{FED}}$ (FED): Multi-round distributed optimization ($T \in \{1, 5, 10, 20\}$ communication rounds) with local EM steps and Hungarian alignment to the global model before server averaging.

\item Transportation medoid $\bar\bstheta^{MED}$ (MED): Selects the single best local density minimizing total transportation divergence to all other local models: $\arg\min_{g\in \{\widehat f_1,\ldots,\widehat f_M\}} \sum_{m=1}^M \lambda_m \Tau_{c}(\widehat f_m,g)$.

\item Naive weighted average $\bar\bstheta^W$ (W-AVR): Direct arithmetic weighted mean of unaligned local parameters $\sum_{m=1}^M \lambda_m\widehat\bstheta_m$, which fails completely due to label switching.

\end{itemize}

The metrics used in experimental studies:
\begin{itemize}[label=--]
{
\item $\mathcal T_c(\cdot, f^*)$: expected transportation divergence of the estimated MoE model to the ground truth $f^*$.

\item $\mathcal T_c(\cdot, \bar f^W)$: expected transportation divergence of the estimated MoE model to $\bar f^W$.

\item Test loglik: log-likelihood of the estimated parameter evaluated on the testing set.

\item $\mathrm{MSE}(\cdot, \bstheta^*)$: mean squared error across all model parameters, computed after matching the estimated experts to the true ones (since the experts are only identified up to a permutation) via searching over all $K!$ permutations.
}

\item RPE: Relative prediction error in test,
\(
\mathrm{RPE} =
\sum_{i=1}^{n}(y_i-\widehat y_i)^2\big/\sum_{i=1}^{n}(y_i-\bar y)^2,
\) where $y_i$ and $\widehat y_i$ denote the true and predicted responses and $\bar y$ their average on the test set. The responses are centred, so that the criterion does not depend on the origin of the response scale.

{
\item ARI: Adjusted Rand Index between the true partition and the clustering predicted by assigning each test observation to the expert with the highest posterior probability (MAP rule).

\item Time (s): Learning time in second required to obtain the estimator. For distributed estimators, it is the largest local fitting time (across  all workers that plays as separate machines) plus the aggregation time.
}



\end{itemize}

\subsection{Data generating process}
\label{sec:data_generating_process}

We evaluate the empirical performance of the estimators under a controlled synthetic setting with $K = 5$ experts and covariate dimension $d = 20$. To investigate sample scalability, we consider total sample size $N = 100\,000$. For each experiment settings (see Appendix \ref{Experimental settings}), we generate 50 independent Monte Carlo replicates.

Each dataset $\sD = \{(\bx_i, y_i)\}_{i=1}^N$ is constructed via a two-stage hierarchical process. First, the covariates are drawn from a Gaussian mixture with $K$ balanced components. Specifically, for each component $k \in [K]$, we draw covariates from $\mathcal{N}_d(\boldsymbol{\mu}_k, \boldsymbol{\Sigma})$, where the cluster centers $\boldsymbol{\mu}_k \in \mathbb{R}^d$ have entries sampled independently from the discrete uniform distribution $\mathcal{U}\{-2, 2\}$. The component covariance matrix $\boldsymbol{\Sigma} \in \mathbb{R}^{d \times d}$ is structured with an autoregressive correlation profile $\Sigma_{uv} = (1/4)^{|u - v|}$ for $u, v \in [d]$. The pooled covariate matrix is subsequently randomly permuted and column-standardized to zero mean and unit variance.

Conditional on the standardized covariates $\bx_i$, the response variables $y_i$ are generated from the conditional MoE density:
\begin{equation*}
y_i \mid Z_i = k, \bx_i \sim \mathcal{N}\big(\beta_{0,k}^* + \bx_i^\top \bsbeta_k^*,\, {\sigma_k^*}^2\big),
\end{equation*}
where the latent expert indicator $Z_i \in \{1, \ldots, K\}$ follows a categorical distribution:
\begin{equation*}
Z_i \mid \bx_i \sim \mathcal{M}\big(1,\, (\pi_1(\bx_i; \bsalpha^*), \ldots, \pi_K(\bx_i; \bsalpha^*))\big),
\end{equation*}
with softmax gating probabilities
\begin{equation*}
\pi_k(\bx_i; \bsalpha^*) = \frac{\exp(\alpha_{0,k}^* + \bx_i^\top \bsalpha_k^*)}{\sum_{j=1}^K \exp(\alpha_{0,j}^* + \bx_i^\top \bsalpha_j^*)}.
\end{equation*}
To ensure parameter identifiability, the gating parameters of the reference expert $K$ are anchored to zero ($\alpha_{0,K}^* = 0$, $\bsalpha_K^* = \mathbf{0}_d$). The remaining gating parameters are sampled from $\mathcal{N}(0, 0.45^2)$ to maintain balanced cluster assignments without gating saturation. The regression intercepts are set to $\beta_{0,k}^* = 0$ for all $k \in [K]$ to compel expert separation strictly through the vectors $\bsbeta_k^* \in \{-10, \ldots, 10\}^d$, thereby inducing label-switching across local estimators. The noise variances are set to ${\sigma_k^*}^2 = 3.0$.

Each generated dataset is partitioned into an $80\%$--$20\%$ train--test split. The distributed and centralized estimators are fitted exclusively on the training sets, while all statistical metrics (except wall-clock runtime) are evaluated on the held-out test sets.

\begin{figure}[!h]
\centering
\subfloat{\includegraphics[width=.5\linewidth]{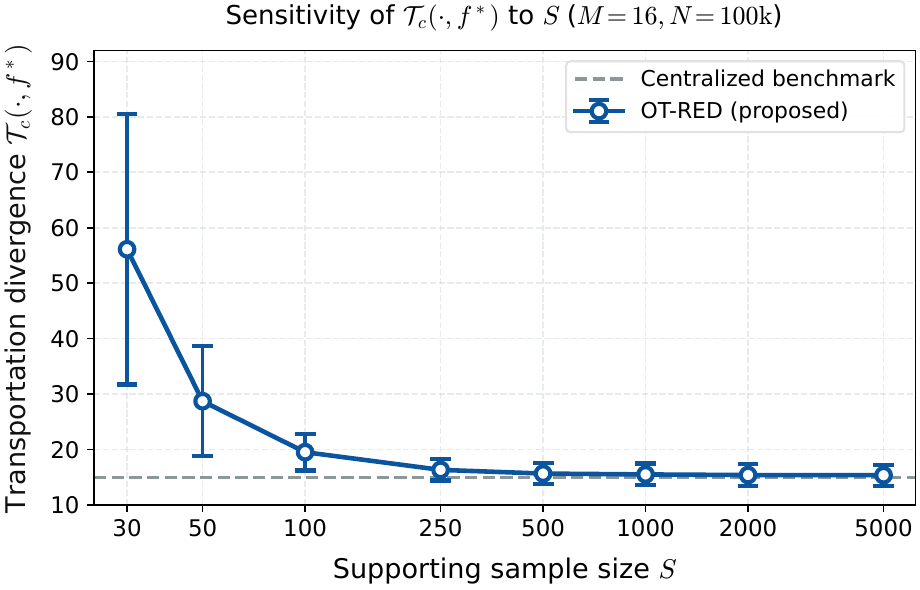}}
\caption{Transportation divergence against supporting sample size $S$}
\label{Figure: sensitivity}
\end{figure}

\subsection{Experimental settings and detailed results}\label{Experimental settings}

\noindent\textbf{Experiment 1: Homogeneous setting (i.i.d. machines).}
Observations are generated from the true global MoE under three mixing proportion regimes: balanced ($\pi_k = 20\%$), moderately imbalanced, and severely imbalanced ($\pi_{\min} = 2\%$). The training data are randomly distributed across the $M=16$ machines so that each machine observes all components and fits a local MoE with $K_m = K = 5$. This experiment examines whether one-shot optimal-transport aggregation can approach the centralized benchmark (GLB) while avoiding the label-switching failures of naive weighted average (W-AVR) and the divergence of multi-round federated averaging (FED).

\begin{table*}[t]
\centering
\caption{Performances of estimators under homogeneous expert capacities ($M=16, N=100k, d=20, K=5$, mean $\pm$ std over 50 runs).}
\label{tab:official_homogeneous}
\setlength{\tabcolsep}{5pt}
\resizebox{\textwidth}{!}{
\begin{tabular}{llccccccc}
\toprule
\textbf{Proportions} & \textbf{Estimator} & $\mathcal{T}_c(\cdot,f^*)$ & $\mathcal{T}_c(\cdot,\bar f^W)$ & \textbf{Test Loglik} & $\textbf{MSE}(\cdot, \bstheta^*)$ & \textbf{RPE} & \textbf{ARI} & \textbf{Time (s)} \\
\midrule
\multirow{10}{*}{\shortstack[l]{\textbf{Balanced}\\\footnotesize $\pi_k = 0.20$}} & GLB & $14.65 \pm 1.90$ & $226.41 \pm 7.03$ & $-52176.7 \pm 152.9$ & $0.0003 \pm 0.0001$ & $0.3885 \pm 0.0076$ & $0.9291 \pm 0.0033$ & $2612.083 \pm 237.065$ \\
 & OT-RED (ours) & $\mathbf{15.04 \pm 2.08}$ & $\mathbf{226.10 \pm 7.04}$ & $-52207.3 \pm 145.4$ & $0.0012 \pm 0.0003$ & $0.3891 \pm 0.0074$ & $0.9289 \pm 0.0031$ & $14.608 \pm 2.572$ \\
 & W-AVR & $234129.9 \pm 5247.6$ & $1.41 \times 10^{6}$ & $-1.47 \times 10^{6}$ & $12.6341 \pm 0.8846$ & $0.8060 \pm 0.0139$ & $0.1146 \pm 0.0172$ & $12.137 \pm 1.988$ \\
 & A-AVR & $17.38 \pm 2.69$ & $228.41 \pm 7.18$ & $\mathbf{-52181.4 \pm 155.5}$ & $\mathbf{0.0005 \pm 0.0001}$ & $\mathbf{0.3885 \pm 0.0077}$ & $\mathbf{0.9291 \pm 0.0033}$ & $12.166 \pm 2.014$ \\
 & GM & $15.17 \pm 2.08$ & $226.27 \pm 7.05$ & $-52206.8 \pm 145.2$ & $0.0012 \pm 0.0003$ & $0.3891 \pm 0.0074$ & $0.9289 \pm 0.0031$ & $12.883 \pm 2.331$ \\
 & MED & $306.61 \pm 69.89$ & $459.04 \pm 31.09$ & $-95360.9 \pm 15541.4$ & $0.2748 \pm 0.1260$ & $1.3159 \pm 0.3571$ & $0.6939 \pm 0.0860$ & $14.839 \pm 3.966$ \\
 & FED ($T=1$) & $26970.2 \pm 108779.1$ & $1222.66 \pm 3645.07$ & $-53233.2 \pm 3908.4$ & $34.5687 \pm 145.9056$ & $0.3935 \pm 0.0245$ & $0.9117 \pm 0.0675$ & $\mathbf{1.728 \pm 0.343}$ \\
 & FED ($T=5$) & $10377.9 \pm 73260.9$ & $557.82 \pm 2329.40$ & $-52510.0 \pm 2332.0$ & $10.2125 \pm 72.2100$ & $0.3906 \pm 0.0190$ & $0.9234 \pm 0.0402$ & $3.791 \pm 0.690$ \\
 & FED ($T=10$) & $10386.6 \pm 73322.4$ & $557.85 \pm 2329.65$ & $-52510.2 \pm 2334.0$ & $10.2195 \pm 72.2597$ & $0.3906 \pm 0.0190$ & $0.9236 \pm 0.0390$ & $10.999 \pm 16.144$ \\
 & FED ($T=20$) & $10383.9 \pm 73303.0$ & $556.92 \pm 2323.06$ & $-52510.7 \pm 2337.1$ & $10.2233 \pm 72.2861$ & $0.3906 \pm 0.0190$ & $0.9241 \pm 0.0351$ & $44.348 \pm 37.335$ \\
\midrule
\multirow{10}{*}{\shortstack[l]{\textbf{Moderately}\\\textbf{imbalanced}\\\footnotesize $0.10 \le \pi_k \le 0.30$}} & GLB & $14.97 \pm 1.83$ & $228.31 \pm 8.51$ & $-51595.8 \pm 122.2$ & $0.0003 \pm 0.0001$ & $0.3662 \pm 0.0061$ & $0.9326 \pm 0.0028$ & $2795.946 \pm 262.610$ \\
 & OT-RED (ours) & $\mathbf{15.57 \pm 1.80}$ & $\mathbf{227.95 \pm 8.50}$ & $-51629.9 \pm 120.4$ & $0.0012 \pm 0.0002$ & $0.3669 \pm 0.0060$ & $\mathbf{0.9326 \pm 0.0026}$ & $17.269 \pm 3.189$ \\
 & W-AVR & $230957.9 \pm 5798.7$ & $1.36 \times 10^{6}$ & $-1.42 \times 10^{6}$ & $12.7092 \pm 0.8983$ & $0.7954 \pm 0.0248$ & $0.1156 \pm 0.0262$ & $15.039 \pm 2.591$ \\
 & A-AVR & $18.21 \pm 2.27$ & $230.75 \pm 8.71$ & $\mathbf{-51601.2 \pm 123.7}$ & $\mathbf{0.0006 \pm 0.0001}$ & $\mathbf{0.3662 \pm 0.0061}$ & $0.9325 \pm 0.0027$ & $15.063 \pm 2.588$ \\
 & GM & $15.77 \pm 1.80$ & $228.17 \pm 8.50$ & $-51629.4 \pm 120.0$ & $0.0012 \pm 0.0002$ & $0.3669 \pm 0.0060$ & $\mathbf{0.9326 \pm 0.0026}$ & $15.724 \pm 2.969$ \\
 & MED & $325.29 \pm 65.67$ & $458.40 \pm 38.54$ & $-99821.7 \pm 16049.2$ & $0.2838 \pm 0.1058$ & $1.4768 \pm 0.3876$ & $0.6815 \pm 0.0836$ & $18.607 \pm 5.246$ \\
 & FED ($T=1$) & $45456.5 \pm 133512.9$ & $2065.05 \pm 4781.28$ & $-53459.3 \pm 4900.0$ & $88.9199 \pm 270.4843$ & $0.3734 \pm 0.0202$ & $0.9142 \pm 0.0548$ & $\mathbf{1.693 \pm 0.358}$ \\
 & FED ($T=5$) & $1.03 \times 10^{8}$ & $2403.12 \pm 10725.17$ & $-63013.9 \pm 75883.0$ & $6.49 \times 10^{14}$ & $12.1586 \pm 83.3554$ & $0.9061 \pm 0.1296$ & $4.316 \pm 2.855$ \\
 & FED ($T=10$) & $3.81 \times 10^{20}$ & $3908.99 \pm 21188.32$ & $-55081.3 \pm 19919.7$ & $3.27 \times 10^{38}$ & $1.10 \times 10^{13}$ & $0.9053 \pm 0.1347$ & $17.560 \pm 24.259$ \\
 & FED ($T=20$) & $25363.4 \pm 101348.5$ & $924.08 \pm 2779.15$ & $-52273.3 \pm 2651.7$ & $52.1899 \pm 210.9950$ & $0.3704 \pm 0.0159$ & $0.9242 \pm 0.0342$ & $50.606 \pm 46.231$ \\
\midrule
\multirow{10}{*}{\shortstack[l]{\textbf{Highly}\\\textbf{imbalanced}\\\footnotesize $\pi_{\min} = 0.02$}} & GLB & $14.83 \pm 2.08$ & $661.59 \pm 1420.20$ & $-49840.9 \pm 175.7$ & $0.0008 \pm 0.0001$ & $0.3009 \pm 0.0072$ & $0.9425 \pm 0.0023$ & $3028.640 \pm 368.578$ \\
 & OT-RED (ours) & $1028.8 \pm 5141.1$ & $\mathbf{356.89 \pm 374.05}$ & $\mathbf{-49964.6 \pm 317.8}$ & $19.3430 \pm 134.9592$ & $0.3020 \pm 0.0071$ & $\mathbf{0.9419 \pm 0.0021}$ & $24.102 \pm 4.398$ \\
 & W-AVR & $233695.9 \pm 8020.8$ & $1.12 \times 10^{6}$ & $-1.18 \times 10^{6}$ & $11.5282 \pm 1.3025$ & $0.7388 \pm 0.0453$ & $0.0979 \pm 0.0494$ & $20.512 \pm 3.460$ \\
 & A-AVR & $\mathbf{293.51 \pm 929.17}$ & $520.60 \pm 968.99$ & $-50010.8 \pm 595.8$ & $\mathbf{0.0938 \pm 0.3246}$ & $\mathbf{0.3011 \pm 0.0074}$ & $0.9419 \pm 0.0028$ & $20.555 \pm 3.493$ \\
 & GM & $3641.8 \pm 17302.6$ & $553.78 \pm 1480.50$ & $-50201.1 \pm 1655.9$ & $7.5943 \pm 27.9151$ & $0.3024 \pm 0.0086$ & $0.9410 \pm 0.0073$ & $21.251 \pm 4.018$ \\
 & MED & $1098.0 \pm 719.7$ & $975.27 \pm 1529.65$ & $-106741.2 \pm 20022.2$ & $0.4152 \pm 0.2086$ & $1.3453 \pm 0.3781$ & $0.6687 \pm 0.1018$ & $22.831 \pm 5.305$ \\
 & FED ($T=1$) & $27446.3 \pm 40359.1$ & $2113.66 \pm 2193.62$ & $-51350.8 \pm 2022.8$ & $93.4367 \pm 144.2760$ & $0.3048 \pm 0.0108$ & $0.9275 \pm 0.0299$ & $\mathbf{1.910 \pm 0.595}$ \\
 & FED ($T=5$) & $4.97 \times 10^{19}$ & $13787.87 \pm 45147.03$ & $-163089.4 \pm 587635.6$ & $1.96 \times 10^{40}$ & $1.72 \times 10^{13}$ & $0.8423 \pm 0.2674$ & $5.369 \pm 3.128$ \\
 & FED ($T=10$) & $7.33 \times 10^{23}$ & $12710.38 \pm 66681.72$ & $-123099.4 \pm 500622.2$ & $4.00 \times 10^{56}$ & $2.23 \times 10^{17}$ & $0.8829 \pm 0.1902$ & $20.529 \pm 36.883$ \\
 & FED ($T=20$) & $1.54 \times 10^{49}$ & $6566.42 \pm 21821.22$ & $-86333.1 \pm 238425.0$ & $3.49 \times 10^{116}$ & $2.91 \times 10^{33}$ & $0.8762 \pm 0.2001$ & $54.189 \pm 63.785$ \\
\bottomrule
\end{tabular}
}
\end{table*}

Table~\ref{tab:official_homogeneous} reports the evaluation metrics for Experiment~1 across the three subpopulation profiles. In the balanced regime ($\pi_k = 20\%$), the proposed OT-RED achieves the lowest transportation divergence to the empirical pooled density $\bar f^W$ ($\mathcal{T}_c = 226.10$) and closely matches the centralized benchmark GLB in divergence to ground truth ($15.18$ vs. $14.73$). Greedy merging (GM) yields comparable performance, while aligned averaging (A-AVR) attains slightly lower parameter MSE at the expense of higher transport divergence ($17.38$). Unaligned averaging (W-AVR) fails catastrophically across all profiles due to label switching, resulting in severe degradation of test log-likelihood and an ARI near zero ($0.1146$).
The transportation medoid (MED) preserves unimodal stability but incurs substantial prediction errors (RPE $> 1.3$) because discarding $M-1$ local models discards most of the sample information.

A crucial finding emerges under subpopulation imbalance. As component rarity intensifies ($\pi_{\min} = 2\%$), multi-round federated averaging (FED) exhibits severe numerical instability: beyond $T=1$ round, local client drift on minority components causes parameter updates to diverge, producing erratic MSEs ($10^{40}$ to $10^{116}$).
One-shot optimal-transport aggregation remains usable in this regime: OT-RED attains the highest test log-likelihood ($-49964.6$), the lowest divergence to $\bar f^W$ ($356.89$), and a high clustering ARI ($0.9419$), slightly below the centralized estimator ($0.9425$). Prediction and clustering are therefore stable, whereas the recovery of the density and of the parameters becomes much more variable for the rarest component, as the standard deviations of $\mathcal T_c(\cdot,f^*)$ and $\Vert\tilde\bstheta-\bstheta^*\Vert_2$ in Table~\ref{tab:official_homogeneous} show.
Across the three regimes, OT-RED runs in $14.6$ to $24.1$ seconds, against $2612$ to $3029$ seconds for centralized training, that is a speedup between $110\times$ and $180\times$.

\noindent\textbf{Experiment 2: Heterogeneous setting (Non-i.i.d. machines).}
To reflect realistic non-i.i.d. environments where local datasets lack certain subpopulations, each machine $m$ observes only a subset of $K_m = 3$ experts (out of $K = 5$), assigned cyclically across machines. Consequently, each worker fits a model with $K_m = 3$, producing a total of $L = 16 \times 3 = 48$ local components sent to the server to be merged into $K = 5$ target experts. In this regime, parameter-matching baselines (A-AVR, W-AVR, MED, FED) are structurally inapplicable due to dimension mismatch ($K_m \neq K$), highlighting the flexibility of OT in aggregating uncoordinated local architectures.

Table~\ref{tab:official_heterogeneous} presents the results of Experiment~2 under uncoordinated local capacities ($K_m = 3 \to K = 5$). Parameter-averaging baselines (A-AVR, W-AVR, FED, MED) are strictly incompatible with this scenario due to the dimension mismatch between local and global parameter spaces. Conversely, optimal-transport reduction naturally accommodates the $L = 48$ local components without structural modification.
OT-RED minimizes the transportation divergence to the pooled model, reaching $\mathcal{T}_c(\cdot, \bar f^W) = 104.13$, below GM ($104.71$) and the centralized model ($105.31$); this is the criterion it optimizes.
In terms of predictive metrics, OT-RED and GM yield the same RPE ($0.4834$) and ARI ($0.9155$), behind the centralized benchmark (RPE $0.3917$) because no machine ever observes the complete five-expert partition. OT-RED obtains this solution in $30.7$ seconds, against $4255.7$ seconds for centralized EM.

\noindent\textbf{Experiment 3: Sensitivity to supporting sample's size.}
Using the homogeneous benchmark data, we examine the empirical sample complexity of the transport objective by varying $S = |\bsX_S| \in \{30, 50, 100, 250, 500, 1000, 2000, 5000\}$. We evaluate three potential sources for $\bsX_S$: pooled covariates across all machines, covariates from a single local machine, and synthetic unlabeled covariates generated from an empirical reference distribution. This assesses the minimal communication overhead required to reliably estimate the gating function.

Table~\ref{tab:sensitivity_S} shows that performance drops sharply when $S < 100$. For $S \in \{30, 50\}$, the transport divergence and prediction error increase substantially ($\mathcal{T}_c(\cdot, f^*) = 91.35$ and $\text{RPE} = 0.5020$ at $S=30$). This is expected: estimating the gating softmax network requires $(d+1)(K-1) = 84$ parameters, leaving the system under-determined for such small sample sizes. 
As soon as $S \ge 100$, however, the estimates stabilize. In fact, $S = 250$ (just $5\%$ of a single worker's dataset) is already enough for practical convergence: OT-RED achieves $\text{ARI} = 0.9319$ and $\text{RPE} = 0.3693$, essentially matching the centralized benchmark ($0.9326$ and $0.3662$) while taking under a second to aggregate ($0.88$~s). Increasing $S$ beyond $500$ brings no noticeable gains in clustering or prediction, but pushes the aggregation time past $31$ seconds at $S = 5000$.
Regarding the choice of $\bsX_S$ (bottom rows of Table~\ref{tab:sensitivity_S}), the results are virtually identical across all three sources. Using covariates from just one worker or generating synthetic samples from a reference distribution yields the same accuracy and divergence as pooling covariates from all machines ($\text{RPE} \approx 0.3665$--$0.3667$, $\mathcal{T}_c(\cdot, \bar f^W) \approx 226.0$). In practice, sending or generating a few hundred unlabeled points ($S \approx 250$) is therefore sufficient, keeping communication negligible without centralizing raw client data.

\begin{table*}[t]
\centering
\caption{Sensitivity of proposed OT-RED to support sample size $S = |\bsX_S|$ and source of $\bsX_S$ ($N=100k, M=16, N_m=5000, d=20, K=5$, mean $\pm$ std over 50 runs).}
\label{tab:sensitivity_S}
\setlength{\tabcolsep}{6pt}
\resizebox{\textwidth}{!}{
\begin{tabular}{lcccccccc}
\toprule
\textbf{Support Condition} & $S/N_m$ & $\mathcal{T}_c(\cdot,f^*)$ & $\mathcal{T}_c(\cdot,\bar f^W)$ & \textbf{Test Loglik} & $\textbf{MSE}(\cdot, \bstheta^*)$ & \textbf{RPE} & \textbf{ARI} & \textbf{Agg. Time (s)} \\
\midrule
$S = 30$ & $0.006$ & $91.35 \pm 104.87$ & $285.55 \pm 87.10$ & $-58103.7 \pm 13249.7$ & $0.0969 \pm 0.2457$ & $0.5020 \pm 0.1527$ & $0.9075 \pm 0.0311$ & $0.433 \pm 0.368$ \\
$S = 50$ & $0.01$ & $33.09 \pm 19.97$ & $240.73 \pm 14.66$ & $-53335.9 \pm 1420.9$ & $0.0234 \pm 0.0243$ & $0.4163 \pm 0.0385$ & $0.9227 \pm 0.0096$ & $\mathbf{0.347 \pm 0.393}$ \\
$S = 100$ & $0.02$ & $19.64 \pm 3.39$ & $230.26 \pm 9.55$ & $-52078.1 \pm 385.3$ & $0.0073 \pm 0.0058$ & $0.3809 \pm 0.0128$ & $0.9299 \pm 0.0033$ & $0.462 \pm 0.553$ \\
$S = 250$ & $0.05$ & $16.48 \pm 2.03$ & $226.81 \pm 9.11$ & $-51706.6 \pm 130.5$ & $0.0023 \pm 0.0012$ & $0.3693 \pm 0.0064$ & $0.9319 \pm 0.0028$ & $0.882 \pm 0.781$ \\
$S = 500$ & $0.1$ & $15.77 \pm 1.99$ & $226.31 \pm 9.13$ & $-51654.6 \pm 126.1$ & $0.0016 \pm 0.0006$ & $0.3675 \pm 0.0059$ & $0.9323 \pm 0.0027$ & $1.401 \pm 1.025$ \\
$S = 1000$ & $0.2$ & $15.60 \pm 1.95$ & $226.11 \pm 9.11$ & $-51635.4 \pm 122.0$ & $0.0013 \pm 0.0003$ & $0.3671 \pm 0.0060$ & $0.9324 \pm 0.0028$ & $3.204 \pm 3.146$ \\
$S = 2000$ & $0.4$ & $15.51 \pm 1.95$ & $226.03 \pm 9.10$ & $-51626.0 \pm 116.9$ & $0.0011 \pm 0.0002$ & $0.3668 \pm 0.0060$ & $0.9325 \pm 0.0027$ & $8.634 \pm 10.776$ \\
$S = 5000$ & $1$ & $15.45 \pm 1.90$ & $\mathbf{225.97 \pm 9.10}$ & $-51622.8 \pm 118.6$ & $0.0011 \pm 0.0002$ & $0.3667 \pm 0.0060$ & $\mathbf{0.9326 \pm 0.0027}$ & $31.564 \pm 23.091$ \\
\midrule
\multicolumn{9}{l}{\textit{Source of $\bsX_S$ ($S = 5000, S/N_m = 1.0$)}} \\
One machine & $1.0$ & $15.48 \pm 1.92$ & $\mathbf{225.97 \pm 9.10}$ & $-51623.2 \pm 118.3$ & $0.0011 \pm 0.0001$ & $0.3667 \pm 0.0060$ & $\mathbf{0.9326 \pm 0.0027}$ & $27.925 \pm 33.743$ \\
Simulated $\mathbf{x}$ & $1.0$ & $\mathbf{15.44 \pm 1.93}$ & $225.99 \pm 9.10$ & $\mathbf{-51611.7 \pm 118.6}$ & $\mathbf{0.0008 \pm 0.0001}$ & $\mathbf{0.3665 \pm 0.0060}$ & $0.9325 \pm 0.0027$ & $15.499 \pm 13.317$ \\
\midrule
\multicolumn{9}{l}{\textit{Reference baselines (independent of $S$)}} \\
GLB (Centralized) & -- & $15.01 \pm 1.87$ & $226.39 \pm 9.11$ & $-51595.9 \pm 122.2$ & $0.0003 \pm 0.0001$ & $0.3662 \pm 0.0061$ & $0.9326 \pm 0.0028$ & $3109.334 \pm 264.719$ \\
A-AVR (Aligned averaging) & -- & $18.11 \pm 2.31$ & $228.78 \pm 9.33$ & $-51601.0 \pm 123.8$ & $0.0005 \pm 0.0001$ & $0.3662 \pm 0.0061$ & $0.9325 \pm 0.0028$ & $0.278 \pm 0.360$ \\
\bottomrule
\end{tabular}
}
\end{table*}

\subsection{Details of applications on real-world data}
\subsubsection{Beijing multi-site air quality}\label{appendix: Beijing multi-site air quality}
\noindent\textbf{Experimental setup.}
To strictly assess operational forecasting without temporal data leakage, we perform an out-of-time evaluation:
\begin{itemize}[leftmargin=*,noitemsep,topsep=1pt]
\item \textit{Training set}: The first three years (March 2013 -- February 2016), containing $N_{\text{train}} = 315\,360$ hourly records partitioned naturally by physical station ($N_m = 26\,280$ per machine).
\item \textit{Testing set}: The entire fourth year (March 2016 -- February 2017), containing $N_{\text{test}} = 105\,120$ out-of-time records ($8\,760$ per station).
\end{itemize}
The task is to predict hourly fine particulate matter ($\text{PM}_{2.5}$ concentration in $\mu\text{g/m}^3$) from $d = 19$ meteorological and chemical covariates (including $\text{SO}_2, \text{NO}_2, \text{CO}, \text{O}_3$, temperature, pressure, dew point, precipitation, wind speed, wind direction, and cyclical temporal indicators). We specify $K = 4$ experts to capture distinct seasonal and atmospheric dispersion regimes (e.g., stagnant inversion layers versus active wind-clearing conditions). Each station fits a local MoE on its own sensor stream. The local models are transmitted to the server and aggregated with $S = 2\,000$ supporting covariates. We benchmark OT-RED against the centralized benchmark (GLB), greedy merging (GM), aligned averaging (A-AVR), naive averaging (W-AVR), and coordinate median (MED). Models are evaluated on the held-out test year using Root Mean Squared Error (RMSE), Mean Absolute Error (MAE), Relative Prediction Error (RPE), Pearson correlation, and total learning time.

\begin{table}[ht]
\centering
\caption{Performance on the Beijing air quality sensor network ($M=12$ stations, $K=4$).}
\label{tab:beijing_air_quality}
{\footnotesize
\setlength{\tabcolsep}{3.5pt}
\begin{tabular}{lccccccc}
\toprule
\textbf{Estimator} & $\mathcal{T}_c(\cdot,\widehat f_{\mathrm{GLB}})$ & $\mathcal{T}_c(\cdot,\bar f^W)$ & \textbf{RMSE} & \textbf{MAE} & \textbf{RPE} & \textbf{Corr.} & \textbf{Time (s)} \\
\midrule
GLB              & $0.00$ & $151.67$ & $17.77$ & $9.46$ & $0.0464$ & $0.9766$ & $29.6$ \\
\midrule
OT-RED (ours)             & $\mathbf{75.60}$ & $\mathbf{112.12}$ & $17.76$ & $9.50$ & $0.0463$ & $0.9766$ & $17.6$ \\
A-AVR            & $118.84$ & $165.71$ & $31.29$ & $11.24$ & $0.1438$ & $0.9308$ & $34.2$ \\
W-AVR            & $1133.92$ & $1158.77$ & $18.30$ & $10.55$ & $0.0492$ & $0.9752$ & $34.2$ \\
GM               & $189.32$ & $261.06$ & $19.39$ & $9.85$ & $0.0552$ & $0.9720$ & $34.2$ \\
MED              & $173.24$ & $153.39$ & $34.70$ & $14.26$ & $0.1769$ & $0.9170$ & $34.3$ \\
\bottomrule
\end{tabular}}
\end{table}

\begin{figure}[ht]
\centering
\subfloat{\includegraphics[width=1\linewidth]{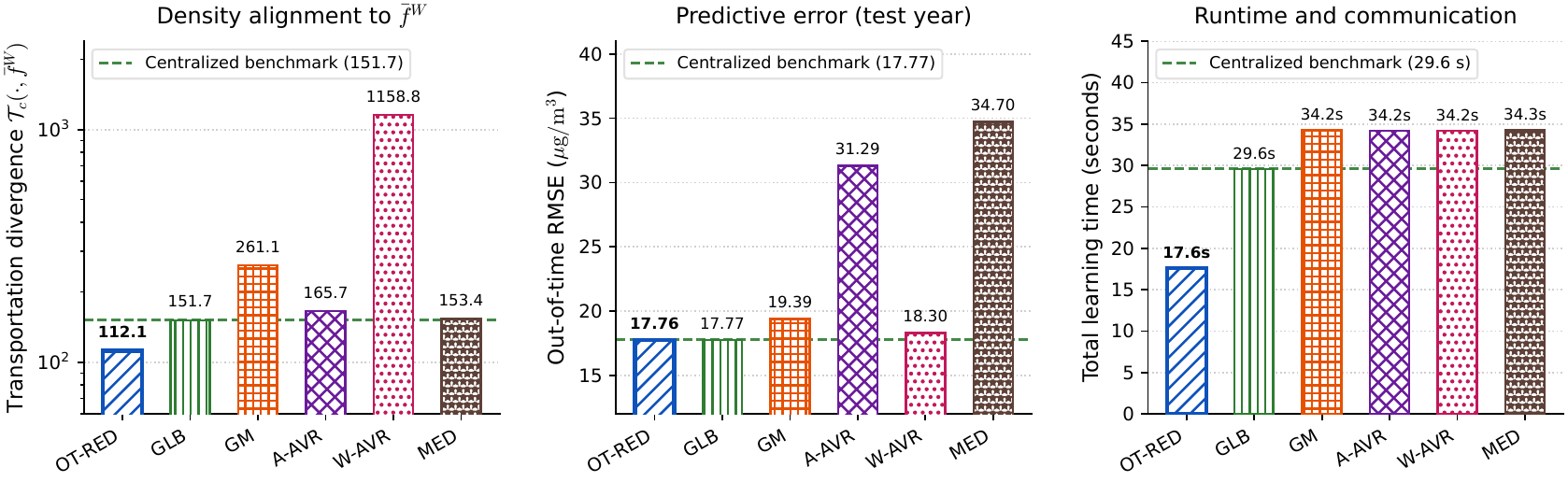}}
\caption{Performance on Beijing multi-site air quality data.}
\label{Figure: figure_beijing_air_quality}
\end{figure}

\begin{figure}[ht]
\centering
\subfloat{\includegraphics[width=1\linewidth]{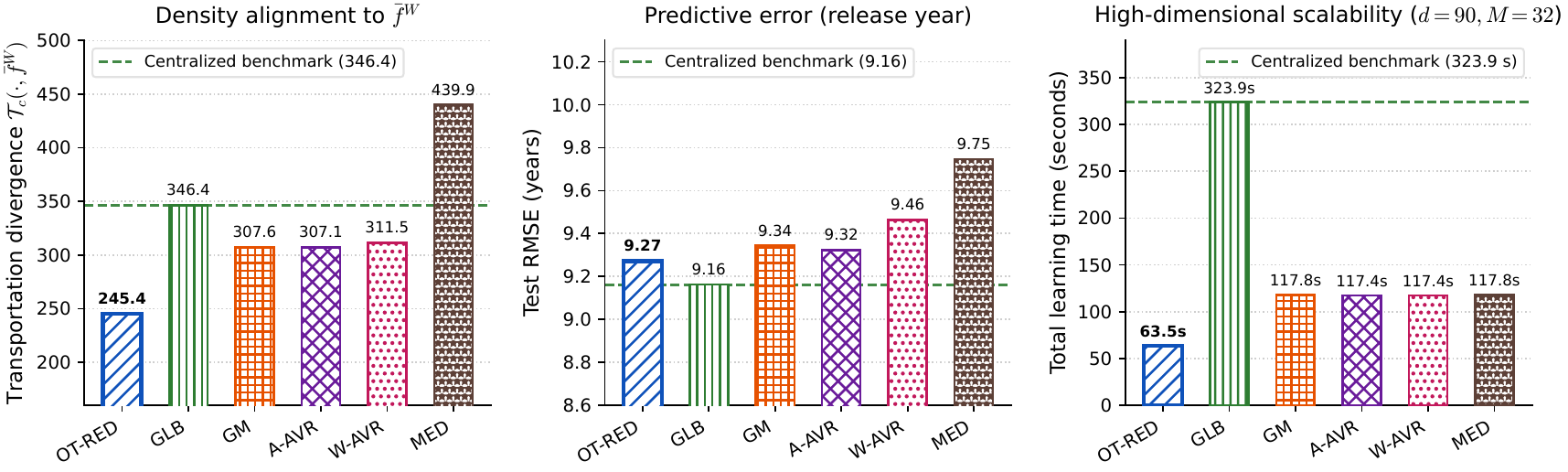}}
\caption{Performance on Year prediction MSD data.}
\label{Figure: figure_year_msd_benchmark}
\end{figure}

\noindent\textbf{Results and discussion.}
Table~\ref{tab:beijing_air_quality} reports out-of-time forecasting performance on the test year. Figure \ref{Figure: figure_beijing_air_quality} plots $\mathcal{T}_c(\cdot,\bar f^W)$, RMSE, and learning time for visualization. OT-RED closely approaches the centralized benchmark GLB across all predictive criteria (RMSE $17.76$ vs.\ $17.77\,\mu\text{g/m}^3$; MAE $9.50$ vs.\ $9.46\,\mu\text{g/m}^3$; RPE $0.0463$ vs.\ $0.0464$; correlation $0.9766$), while reducing learning time from $29.6$ to $17.6$ seconds. Crucially, in distribution space, OT-RED achieves the lowest transportation divergence to the empirical mixture $\bar f^W$ ($\mathcal{T}_c = 112.12$, lower even than GLB at $151.67$) and stays closest to GLB ($\mathcal{T}_c = 75.60$). It is also nearly twice as fast as GM and A-AVR ($34.2$ seconds), while A-AVR suffers substantial degradation (RMSE $31.29$) due to station-specific micro-climatic differences distorting direct parameter averaging.

Unaligned W-AVR and MED perform poorly (RMSE $18.30$ and $34.70\,\mu\text{g/m}^3$). Overall, OT-RED matches the centralized benchmark on the predictive criteria and gives the lowest divergence to the pooled density, the criterion it minimizes, in a single one-way communication round, which suits networks where multi-round coordination is impractical.

\subsubsection{Year prediction MSD}\label{appendix: Year prediction MSD}

As shown in Table~\ref{tab:year_msd_benchmark}, OT-RED closely approaches centralized training (RMSE $9.27$ vs.\ $9.16$ years; MAE $6.46$ vs.\ $6.37$ years; RPE $0.7300$ vs.\ $0.7128$; $\text{Corr}=0.5202$ vs.\ $0.5365$) while running in $63.5$ seconds, compared to $323.9$ seconds for GLB (a $5.1\times$ speedup) and roughly $117$ seconds for GM and A-AVR. In distribution space, OT-RED attains the lowest transportation divergence to the pooled mixture $\bar f^W$ ($\mathcal{T}_c = 245.4$), well below GLB ($346.4$), GM ($307.6$).

Unaligned averaging (W-AVR) and coordinate median (MED) perform poorly (RMSE $9.46$ and $9.75$ years). These results demonstrate that one-shot optimal-transport aggregation scales effectively to high-dimensional settings ($d=90$) and large worker counts ($M=32$). Figure \ref{Figure: figure_beijing_air_quality} plots the $\mathcal{T}_c(\cdot,\bar f^W)$, RMSE, and learning time to visualize the comparison.

\end{document}